%% file: main_arxiv.tex
\documentclass[letterpaper]{article} 
\usepackage{arxiv}
\usepackage[utf8]{inputenc} 
\usepackage[T1]{fontenc}    
\usepackage{url}            
\usepackage{booktabs}       
\usepackage{amsfonts}       
\usepackage{nicefrac}       
\usepackage{microtype}      
\usepackage{graphicx}
\usepackage{natbib}
\usepackage{hyperref}       
\usepackage{doi}
\usepackage{caption}
\usepackage{enumitem}
\usepackage{tcolorbox}
\usepackage{array} 
\newcommand{\anonymize}[2]{#1}

\usepackage{ifthen}
\newboolean{includeappendix}
\setboolean{includeappendix}{false} 

\title{Estimating Contribution Quality in Online Deliberations Using a Large Language Model}
\author{ 
    Lodewijk L. Gelauff $^*$ \\
	Center on Democracy, Development and the Rule of Law\\
	Stanford University\\
	\texttt{lodewijk@stanford.edu} \\
\And 
    Mohak Goyal\thanks{These authors contributed equally.} \\
	Management Science and Engineering\\
	Stanford University\\
	\texttt{mohakg@stanford.edu} \\	
 \And
    Bhargav Dindukurthi\\
    Cadence Design Systems\\
	\texttt{bhargavdindukurthi1@gmail.com} \\
 \And
    Ashish Goel\\
	Management Science and Engineering\\
	Stanford University\\
	\texttt{ashishg@stanford.edu} \\
 \And
    Alice Siu\\
	Center on Democracy, Development and the Rule of Law\\
	Stanford University\\
	\texttt{asiu@stanford.edu} \\
}

\begin{document}
\maketitle

\input{content}

\section*{Acknowledgments}
We thank James Fishkin, Lovish Chopra, and Shoaib Mohammed for their valuable discussions. We are grateful to Tenzin Kartsang for coordinating the human annotators, as well as to the annotators themselves. We also extend our thanks to our deliberation partners, particularly those at Meta, for their assistance in data collection. Additionally, we acknowledge the use of ChatGPT for enhancing the clarity of parts of this paper.

\bibliographystyle{apalike}
\bibliography{refs}

\newpage
\appendix
\input{appendix}

\end{document}

%% file: content.tex
\begin{abstract}

Deliberation involves participants exchanging knowledge, arguments, and perspectives and has been shown to be effective at addressing polarization.
The Stanford Online Deliberation Platform facilitates large-scale deliberations. It enables video-based online discussions on a structured agenda for small groups without requiring human moderators. This paper's data comes from various deliberation events, including one conducted in collaboration with Meta in 32 countries, and another with 38 post-secondary institutions in the US.

Estimating the quality of contributions in a conversation is crucial for assessing feature and intervention impacts. Traditionally, this is done by human annotators, which is time-consuming and costly. We use a large language model (LLM) alongside eight human annotators to rate contributions based on justification, novelty, expansion of the conversation, and potential for further expansion, with scores ranging from 1 to 5. Annotators also provide brief justifications for their ratings. Using the average rating from other human annotators as the ground truth, we find the model outperforms individual human annotators. While pairs of human annotators outperform the model in rating justification and groups of three outperform it on all four metrics, the model remains competitive.

We illustrate the usefulness of the automated quality rating by assessing the effect of nudges on the quality of deliberation. We first observe that individual nudges after prolonged inactivity are highly effective, increasing the likelihood of the individual requesting to speak in the next 30 seconds by 65\%. 
Using our automated quality estimation, we show that the quality ratings for statements prompted by nudging are similar to those made without nudging, signifying that nudging leads to more ideas being generated in the conversation without losing overall quality.
\end{abstract}



\maketitle

\section{Introduction}

Deliberation is a practice that can help bring together stakeholders and ensure an exchange of opinions, facts and understanding. 
Deliberation is used in various contexts, including mini-publics \cite{goodin_deliberative_2006}, citizen juries \cite{smith_citizens_2000} and deliberative polls \cite{fishkin1997voice}, to better understand a population's opinions on a complex topic or inform a decision. Some of these designs have reported promising spillover effects such as depolarization \cite{van_der_does_small-scale_2023, fishkin_is_2021}.

Conducting deliberation online offers numerous advantages. In-person deliberation can be prohibitively expensive due to transportation and venue costs \cite{luskin_online_2014}. Additionally, organizing unbiased and reproducible moderation is challenging and requires extensive training for human moderators. Online deliberation, however, can leverage automated processes that combine group intelligence and artificial intelligence to replace human moderators \cite{gelauff2023achieving}. This not only reduces costs but also ensures consistent and scalable facilitation.


Facilitating online deliberation presents both opportunities for data analysis and the responsibility to ensure the process aligns with its objectives. Key factors to consider include equitable participation, conversation quality, participant experience, and outcome quality. The online setting simplifies data collection and makes evaluation more affordable. As the scale of deliberation increases, rapid evaluation becomes crucial for organizers to adjust recruitment, materials, or process design promptly. In this paper, we design a scalable method for quantifying deliberative quality and show how it can assist platform designers in evaluating and improving moderation interventions and configurations.




Different aspects of quality can be established by analyzing participant surveys, third-party observations, or the content of the deliberation. Especially the content (transcripts) of the deliberation provide a promising avenue to evaluate interventions and moderation decisions that affect the conversation locally. Traditionally, content analysis relies on human annotators to categorize or score contributions -- a resource-intensive method. The arrival of LLMs makes it feasible to use automated analysis to estimate the quality of a contribution, exchange or entire deliberation. We define criteria of interest, and design a query for the LLM to report scores for statements given the preceding contributions and other relevant context. We validate scores from these queries with human annotators. 

This approach not only allows analysis at a much larger scale but also enables the use of such scores in real-time evaluation, information systems for human supervisors, and even moderation interventions. We illustrate one use of this approach by analysing the effect of sending targeted nudges to inactive participants on the quality of the conversation.



\paragraph{Roadmap}
After discussing related work, we will introduce our data, platform, and methodology. We will explain how we use an LLM to estimate the quality of contributions, and how we evaluate and utilize these estimates. Next, we will compare the performance of our model to that of human evaluators. Finally, we will demonstrate how these quality estimations can be used to optimize the performance of the deliberation platform by comparing the quality of statements made after nudges to those made without nudges. 

\section{Related Work}
\label{sec:relatedwork}

Evaluating a deliberation or discourse has long been a question of interest, with the aspects of analysis evolving over time. Qualities often sought after include equality, rational reasoning and engagement \cite{dutwin_character_2003}. \citet{black2014methods} identified various approaches to evaluating a deliberation. Besides indirect measurements (looking at antecedents and outcomes), they distinguish between micro-analytic and macro-analytic approaches. Micro-analytic approaches include content analysis (annotating units of conversation, usually based on a codebook by humans) and discourse analysis (qualitative analysis of the discourse). Macro-analytic approaches use a wider range of methods to gain insights into the overall deliberation event with commonly used self-assessments from participants, moderators and/or observers or case studies with an in-depth evaluation of some of the groups. For our purposes, content analysis is particularly relevant, and we will focus on that. 

The criteria used in content analysis are not universally agreed upon and depend on what the evaluator is trying to understand. For example, \citet{gerhards_diskursive_1997} looked at the abortion debate in major German newspapers between 1970 and 1994, and evaluated representativeness, respect, level of justification and rationality. \citet{steenbergen_measuring_2003} looked at a parliamentary setting with (long, formal) spoken contributions. They evaluated them on a \textit{Discourse Quality Index} with indicators of openness of participation, justification of demands, common good/empathy, respect (for the group, the topic and arguments) and constructiveness. In line with this work, \citet{gold_visual_2015} defined four indicators and combined the content analysis with a visualization, defining participation (distribution of contributions), respect (frequency of interruptions), argumentation (looking at sentence structure) and persuasiveness (as an indirect metric).

In online education, a lot of work discusses how to evaluate individual contributions to evaluate learning. While this is a slightly different setting, the considered criteria overlap with those of deliberation. \citet{mason_evaluation_1992} proposed a list of six questions that an educational analyst could be asked in content analysis to evaluate the quality of student learning: 

\begin{itemize}
    \item do the participants build on previous messages?
    \item do they draw on their own experience? 
    \item do they refer to course material? 
    \item do they refer to relevant material outside the course? 
    \item do they initiate new ideas for discussion? 
    \item does the course tutor control, direct or facilitate?
\end{itemize}

\citet{henri_computer_1992} proposed an analytical framework with five dimensions (participative, social, interactive, cognitive and metacognitive), of which the cognitive is most relevant to content analysis, with detailed indicators. \citet{newman1995content} built on both works and proposes a set of indicators on relevance, importance, novelty, bringing outside knowledge, ambiguity clarification, linking ideas, justification, critical assessment, practical utility and width of understanding. 

\citet{friess2015systematic} gave an overview of the research on online deliberation platforms.
\cite{esau_design_2017} evaluated the design of (asynchronous) news discussion platforms through content analysis on rationality, reciprocity (engagement), respect and constructiveness. 
We adapt this literature to come up with four representative quality criteria, which we describe in \S\ref{sec:methodology estimating}.

Challenges to content analysis are myriad. 
The manual annotation of statements requires considerable resources, and even then, reasonable and qualified annotators can disagree on quality indicators \cite{black2014methods}. 

LLMs have been a transformative technology \cite{devlin2018bert, vaswani2017attention, brown2020language}, driving significant advancements in natural language understanding and generation. 
Historically, LLMs have been employed in various coding and annotation tasks, consistently outperforming humans in speed and often in quality \cite{goel2023llms, park2024leveraging, refuel_ai}. While most applications focus on factual annotations, we utilize LLMs to tackle the subjective task of assessing the quality of deliberative contributions, a process made more challenging by the lack of an established ground truth.

Several researchers have explored the use of LLMs in democratic processes. \citet{small2023opportunities} discuss integrating LLMs into their text-based deliberation platform, Polis, for tasks such as moderation, summarization, vote prediction, and consensus identification. Other studies, such as those by \cite{bakker2022fine, fish2023generative, ding2023self}, provide fine-tuning approaches to generate consensus statements from text-based opinions. 
Particularly relevant to our work, \citet{fish2023generative} use LLMs as generative and discriminative oracles to develop consensus statements and evaluate participants' utility from it. We use LLMs as oracles for the quality of statements and use them to inform the platform's design.
\citet{li2023prd} demonstrated using LLMs to evaluate subjective contributions made by other LLMs.

\section{Data}
\label{sec:data}
\subsection{Deliberation Platform}
\label{sec:platform}

The data we used for this study was collected from deliberations on the \anonymize{Stanford Online Platform for Deliberation}{Anonymous Deliberation Platform}. This platform is designed to support large-scale online deliberations, with multiple small-group rooms deliberating in parallel on the same agenda, all without the need for a human moderator. The platform mimics the setup of an in-person deliberative poll. Participants are assigned to rooms, each of which is part of a nested structure:
\begin{itemize}[noitemsep]
    \item \textit{Room}: a small group of typically 5--12 participants who discuss with each other. 
    \item \textit{Roomgroup}: a set of rooms deliberating in parallel, using an identical configuration of the deliberation platform, with an identical agenda. 
    \item \textit{Session}: Roomgroups that discuss the same agenda, typically with nearly identical configurations, but not necessarily at the same time. This could, for example, facilitate different availability, privacy preferences or languages. 
    \item \textit{Deliberation event}: a series of sessions where the same group of participants discusses a set of agendas. A deliberation event typically consists of 1--5 sessions.
\end{itemize}

While the platform supports various configurations and exceptions, we will describe a typical deliberation setup. The organizer of the deliberative event recruits a representative sample of their population of interest through polling firms and shares background materials. The agenda has 3-6 agenda items, each with a balanced set of arguments in favor and against the proposals. Participants often receive these materials before the event. 
Participants are then invited with details on the date and time of each session, along with a link to the landing page for their assigned room group. An admin team is available to assist participants with technical issues, supervise discussions, and intervene in case of technical or behavioral problems.

The participants are in advance, at the starting time or on arrival (randomly) distributed across rooms, where the software considers a minimum size and target size for each room. 
When the first small-group session begins, participants in each room are greeted with a video message and a reintroduction to the conversation topic. They watch an instructional video demonstrating the platform's buttons and functions. Participants then introduce themselves, appearing in small video screens with the speaker’s video enlarged.
Speaker turns are managed by an automated moderator using a visible queue that participants can join by requesting to speak. Each participant can speak for up to 45 seconds, after which the next person in the queue takes their turn.

Each agenda item is introduced with an audio or video prompt, accompanied by a description and key arguments for and against the proposals, visible on the screen. Participants can propose to move to the next agenda item if they feel the topic has been sufficiently discussed. If a majority agrees, or if time runs out, the automated moderator introduces the next topic and presents the new arguments.

Once all agenda items have been discussed, the room enters the question development phase. Participants propose questions about the discussed topics and rank them by importance. The group then discusses the top-ranked questions, and the original proposer can edit their question. After discussing all questions or when time expires, the group votes on two questions to send to the panel of experts in the following plenary session.
Participants are then redirected to the plenary session, usually after a short break, where a panel of experts answers their questions. After this, participants may be asked to join a second session, where they might be randomly reassigned to different groups.

\subsection{Deliberation Events}
\begin{table*}[h]
    \centering
    \begin{tabular}{p{0.4cm}p{4.5cm}p{1cm}p{1.3cm}p{1.3cm}p{1.2cm}p{1cm}p{1cm}p{1cm}} 
        \hline
        ID & Topics & Rooms & Unique Participants & Filtered Contributions & Sessions & Median Room Size & Mean Room Size & Average Length (Characters)\\
        \hline
        E1 & Climate change; voting systems & 116 & 583 & 6232& 2& 9 & 8.66 & 513.64 \\
        E2 & Regulating bullying and harassment in virtual reality (Meta) & 221 & 698 & 7015 &  4& 8 & 8.34 & 386.56 \\
        E3 & Regulating bullying and harassment in virtual reality (Meta) & 47 & 149 & 1824 & 4& 9 & 8.57 & 376.24 \\
        \hline
        E4 & (various, including E2) & 1368 & 4654 & 54990 & - & 8 & 7.68 & 425.31 \\

    \end{tabular}
    \small
    \caption{Sets of rooms used for analysis with general topic and statistics. A participant typically participates in one room per session within the same deliberation event. 
    }
    \label{tab:data}
\end{table*}


The sets E1, E2 and E3 refer to rooms that were part of three deliberation events. E1 (``Shaping Our Future") was an event in May 2021, conducted in collaboration with 38 postsecondary institutions with a representative sample of participants from the United States, discussing climate change in one session and voting systems in another \anonymize{\cite{beyene_shaping_2021}}{(Anonymous et al. 2021)}. E2 and E3 were two sets of roomgroups, part of a larger event in 2022, conducted in collaboration with Meta 
(formerly Facebook), with a sample of participants from North America (E2) and the United Kingdom (E3), both discussing the same agenda on regulating bullying and harassment in virtual reality (the ``Metaverse") \anonymize{\cite{fishkin_public_2023}}{(Anonymous et al. 2023)}. The entire event was held in 32 countries in their respective languages. 
These events were selected because of the availability of matched demographic information. When we were limited in how much data we could analyze, especially involving human annotators, we chose E1 for our analysis, as it used the most representative agenda items.

In the set E4, we consider roomgroups on the platform between July 19th, 2022, and March 26th, 2024 configured with language settings English (United States) and at least five rooms with a transcript. This set includes rooms in the Meta deliberation in E2 and several other events and agendas. We use this set of rooms to study the nudge effectiveness, and only include roomgroups that were active after the introduction of a change in the software where some nudges would be skipped (as explained in \S \ref{sec:methodology nudges}).

We filtered contributions by agenda item (removing any contributions during the introductions phase and the question development phase) and removed statements of less than 100 characters in the transcript (short statements are often less reliable and may include more clarifications and technical queries). We refer to these as filtered contributions. Table~\ref{tab:data} describes some statistics of these roomsets. 


\section{Methodology}
\label{sec:methodology}


We design a prompting scheme for using LLMs to evaluate the quality of statements made in online video-conferencing-based deliberations. We then conduct experiments with human annotators to validate the usefulness of the LLM evaluations. Using these quality evaluations, we study the effect of nudges on our deliberation platform on the quality of the statements made. We use the current state-of-the-art GPT-4 model from OpenAI \cite{achiam2023gpt}, which we will refer to as ``the model". 

\subsection{Estimating the Quality of a Statement}
\label{sec:methodology estimating}
There are many possible definitions for the ``quality'' of a contribution in the context of a deliberation, as discussed through the various examples in \S\ref{sec:relatedwork}. In our study, we build on the goal of a deliberative poll, which is to provide participants with optimal conditions to form an informed opinion on the topic at hand \cite{fishkin2009people}. 

We based three of our criteria on the questions and indicators proposed by Mason and Newman \cite{mason_evaluation_1992, newman1995content}:  whether the contribution \textit{provides justification, is novel, expands the conversation}. We added a fourth criterium whether the contribution \textit{allows further expansion} because we're interested in how contributions in a deliberative conversation fit together. 
Specifically, we ask the annotators to score the statement on following criteria prompts:
\begin{itemize}[noitemsep]
    \item [Q1] This statement includes examples or anecdotes to support the speaker's point. 
    \item [Q2] This statement introduces novel ideas, perspectives, or solutions.
    \item [Q3] This statement builds on top of the previous statements and the proposal.
    \item [Q4] This statement raises points which will likely improve the quality of the following discussion. 
\end{itemize}
In the rest of the paper, we will refer to these prompts as Q1, Q2, Q3, and Q4 for brevity.
%
%
Each time, we provided the annotator with the contribution, the topic of the current agenda item, and all preceding statements within the same agenda item. We then asked the annotator to rate their agreement with Q1--4 on a 5-point Likert scale (Strongly Disagree -- Strongly Agree) and provide a justification for each criterion. 
The exact query given to the model is as follows:
\begin{tcolorbox}[colback=gray!10,colframe=gray!80,title=Query for Estimating the Quality of a Statement,
                  boxrule=0.5pt, boxsep=2pt, left=2pt, right=2pt, top=2pt, bottom=2pt]
 \textbf{System Instructions:} Your expertise is required to help a research study to evaluate the quality of statements made in a deliberation on a topic of social interest.

\textbf{Task:} You are observing a live deliberation on the following proposals $<$\textit{topic}$>$. \\
The input is from a noisy speech-to-text system. 
  Your task is to evaluate a statement in the context of the ongoing discussion. 
  Specifically, you have to rate it on the standard Likert scale on 1 to 5 on whether: $<$\textit{criterium statement}$>$.\\
  This is what the rating on the standard Likert scale means. 1: Strongly Disagree. 2: Disagree. 3: Undecided. 4: Agree. 5: Strongly Agree.
  
  Please give a succinct justification in one short sentence. Format your answer as follows: \\
  Rating: x/5. Justification: $[$One short sentence$]$. 

   Here is the transcript of the deliberation from before the statement which is to be evaluated. This serves as the context for the evaluation. $<$\textit{previous contributions}$>$.
   
  Here is the statement which you have to evaluate.  $<$\textit{contribution}$>$.
\end{tcolorbox}

The system instructions aim to improve the quality of responses by \textit{assigning a role} \cite{wu2023large}. 
The query was refined by experimenting with the model's evaluations of ten randomly chosen contributions. This process helped us develop a query that enables the model to consistently distinguish between good and bad statements and provide meaningful justifications for its ratings.

\subsection{Evaluating Automated Quality Estimations}
\label{sec:methodology evaluating}
In order to evaluate the quality estimates created by our quality model, we need to compare these estimates with a baseline. If we were presented with the challenge to estimate the quality of a contribution without access to language models, we would ask one or some research assistants (often familiar with the deliberation procedures to score them: interns, undergraduate and graduate students) to score the contributions, and if necessary take the average of those scores. For an evaluation, we want to compare the score of a human estimate with the model estimate.

For this study, we recruited eight research assistants associated with our lab, but not directly involved with this project. 
We identified a deliberation in English with participants based in the United States where the gender information of participants was readily available (E1 in Table~\ref{tab:data}). These rooms had 6232 filtered contributions, discussing a specific agenda item with a transcript of at least 100 characters. We randomly selected 15 of these statements from male and female participants each.

We asked the human annotators to perform two tasks: first, to estimate the quality of a contribution based on each of the four criteria and justify their score in a sentence; second, to evaluate a quality estimation, which includes a rating and its justification. We divided the human evaluators randomly into two equal cohorts, and split the statements by gender. In the first round, the first cohort estimated the quality of the first set of 15 statements and scored a rating-justification pair for each statement generated by the model but unknown to them. Meanwhile, the second cohort performed the same tasks on the second set of statements.
In the second round, the first cohort estimated the quality of the second set of statements. After each contribution, they scored five different rating-justification pairs (one from the model and four from the other cohort of evaluators, in random order). The second cohort did the same for the first set of statements. For each evaluation of a rating-justification pair, the human evaluators did not know its source.



\subsection{Nudging to Encourage Participation}
\label{sec:methodology nudges}
The platform implements nudges as pop-up overlays on the user's screen. The pop-up includes a phrase encouraging the user to contribute to the conversation or asking whether the group is ready to move on to the next agenda item. In this paper, we will focus on the nudges that encourage users to participate (a ``Speak'' nudge) when they have not participated for a period of three minutes. The textual content of the nudge is randomly drawn from a list of possible nudges. This list varies between events and can be broadly categorized in a \textit{general nudge} (e.g. ``What would someone who disagrees with you say about this discussion''), a \textit{personalized nudge} that addresses the user by their screen name (e.g. ``XYZ, you haven't spoken in a while, is there something you would like to share with the group?'') or a \textit{procon nudge} that explicitly refers to the listed argument in the agenda item that is estimated to have been discussed least. Since 2022, the platform has also included dummy nudges for research purposes: the platform would randomly ``skip" a speak nudge altogether (typically 15\%). 
The platform also emits Speak-Room nudges to all participants when no one has spoken for over 30 seconds. We do not study the effect of Speak-Room nudges in this work.

\section{Evaluating Automated Quality Estimation}
\label{sec:performance}


We will first establish the level of consensus between the human annotators (\S~\ref{sec:methodology evaluating}), and then use this as a benchmark for the model's performance. 


\subsection{Inter-Rater Agreement of Human Annotators}
Inter-rater reliability (IRR) or inter-rater agreement measures the consensus among different annotators of a dataset. 
A common statistic for studying IRR is within-group reliability ($r_{wg}$) \cite{james1993rwg}. While there are multiple variants of $r_{wg}$ \cite{o2017overview}, we report $r^*_{wg}$ as it is best suited for our setting where multiple annotators report Likert scale scores for many statements. 
\begin{equation}\label{eq:rwg_star}
    r^*_{wg} = 1 -  \frac{S_x^2}{\sigma^2_{eu}}.
\end{equation}

Higher $r^*_{wg}$ indicates better IRR. Here $S_x^2$ is the mean of the observed variance in
annotators’ ratings on each item, and $\sigma^2_{eu}$ is the variance of the rectangular, uniform null distribution,
$(m^2- 1)/12$, where m is the number of discrete Likert-type response options. Here $m = 5.$


The $r^*_{wg}$ values for the human annotators are as follows: Q1: 0.519; Q2: 0.449; Q3: 0.605, and Q4: 0.496.
Although interpreting these values can vary by field, we follow the standards recommended by \citet{lebreton2008answers}: 0–0.30 (lack of agreement), 0.31–0.50 (weak agreement), 0.51–0.70 (moderate agreement), 0.71–0.90 (strong agreement), and 0.91–1.0 (very strong agreement). The human annotators show weak agreement on novelty (Q2) and their potential for further expansion (Q4). They have moderate agreement on whether contributions justify their claims (Q1) and expand the conversation (Q3).

\begin{table}[h]
    \centering
    \begin{tabular}{p{0.4cm}>{\centering\arraybackslash}m{0.8cm} >{\centering\arraybackslash}m{1cm} >{\centering\arraybackslash}m{1.5cm} >{\centering\arraybackslash}m{2.4cm}}
        \toprule
        Q & Model Mean & Humans Mean  & Mean Difference & Mean Difference 95\%  CI \\
        \midrule
        Q1 & 2.50 & 3.14 & -0.64 & [ -0.98, -0.34] \\
        Q2 & 2.93 & 3.35 & -0.42 & [ -0.67, -0.15] \\
        Q3 & 4.03 & 3.79 & 0.24 & [0.05, 0.42] \\
        Q4 & 4.10 & 3.79 & 0.31 & [0.04, 0.54] \\
        \bottomrule
    \end{tabular}
    \caption{Average scores given by the model and humans with 95\% confidence intervals for the difference.}  \label{tab:scores}
\end{table}
\begin{table}[h]
    \centering
    \begin{tabular}{p{0.5cm} >{\centering\arraybackslash}m{3cm} >{\centering\arraybackslash}m{3cm}}
        \toprule
        Q & Difference (Men) & Difference (Women) \\
        \midrule
        Q1 & -0.77 & -0.51 \\
        Q2 & -0.62 & -0.21 \\
        Q3 & 0.14  & 0.34  \\
        Q4 & 0.44  & 0.18  \\
        \bottomrule
    \end{tabular}
    \caption{Difference in average scores given by the model and humans for both genders. A negative value means that the model gives lower scores than humans.}
    \label{tab:scores-gender}
\end{table}
\subsection{Ratings from Humans and the Model}
We report the humans' and model's average scores across the 30 statements in Table~\ref{tab:scores}.
\ifthenelse{\boolean{includeappendix}}{
   See Fig.~\ref{fig:scores-bar} in Appendix~\ref{app:scores} for the score distributions.}{} 
We obtained a 95\% confidence interval for the difference of the means by taking 10,000 bootstrap resamples of the original data and calculating the mean difference for each resample. 
We observe that the model consistently gives lower scores than humans on Q1 and Q2 and slightly higher scores on Questions 3 and 4. Addressing this bias -- whether through prompt adjustments, fine-tuning, or mean-correcting the model's scores -- is an important direction for future research. In this study, we use the raw scores. We suggest a mean-correction approach in 
\ifthenelse{\boolean{includeappendix}}{
   Appendix~\ref{appendix-mean correction}, which shows that the model's performance can be further enhanced via this simple correction.}{the full version of the paper, which shows that the model's performance can be enhanced via this simple correction.} 

The bias in the model's scores is more or less consistent across the genders of participants, as in Table~\ref{tab:scores-gender}.
However, since our data is over only 15 statements for men and women each, further investigation is required to ensure that the model is not biased.

A common approach to improve noisy ratings is to take the average of multiple individual ratings. We will compare the model's rating against ratings of individual annotators, as well as pair and triplet ratings (defined by the average of its members' ratings). The average rating of the remaining human annotators serves as the ``golden" rating. Both the model and the human groups are evaluated on how closely their ratings match this golden rating\footnote{We do not consider the golden rating to be the average of \textit{all} human annotators since that will give an unfair advantage to the group of humans since their ratings will be included in the average.}.

\begin{figure}[h]
    \centering
   \includegraphics[trim=2pt 2pt 5pt 5pt, clip, scale = 0.55]{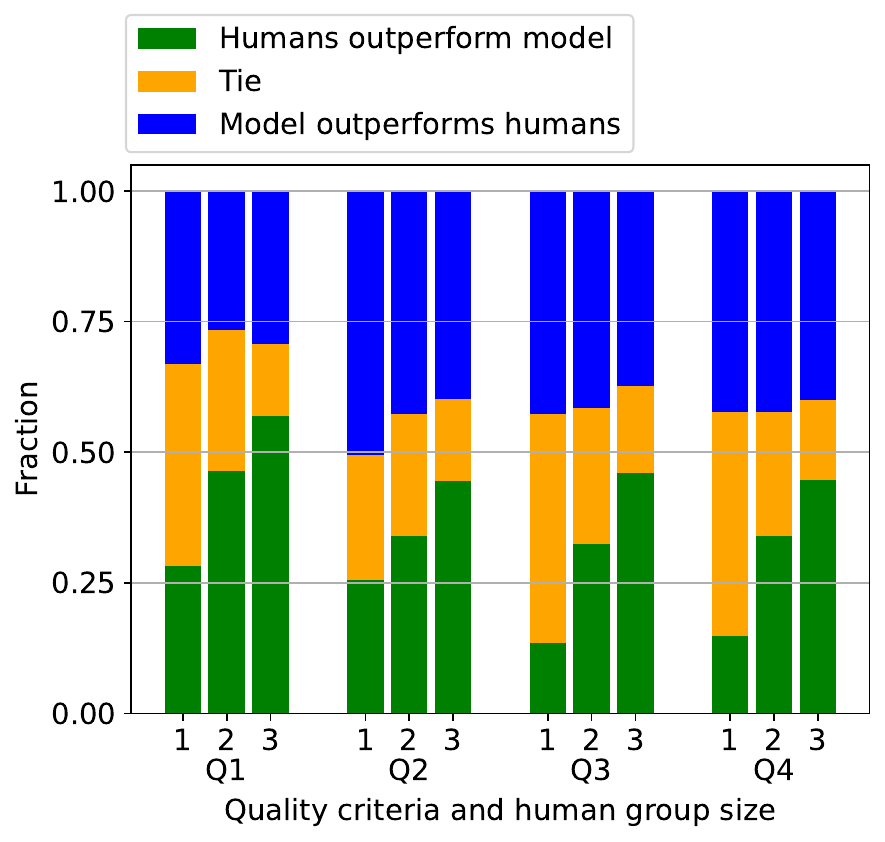}
    \caption{Fraction of data points where groups of humans are outperformed by the model. The model `wins' if it rates a statement closer to the golden rating than a group of humans.}
\label{fig:humans-vs-model-8cg-times-30}
\end{figure}

\begin{figure}[h]
    \centering
   \includegraphics[trim=2pt 2pt 5pt 5pt, clip, scale = 0.55]{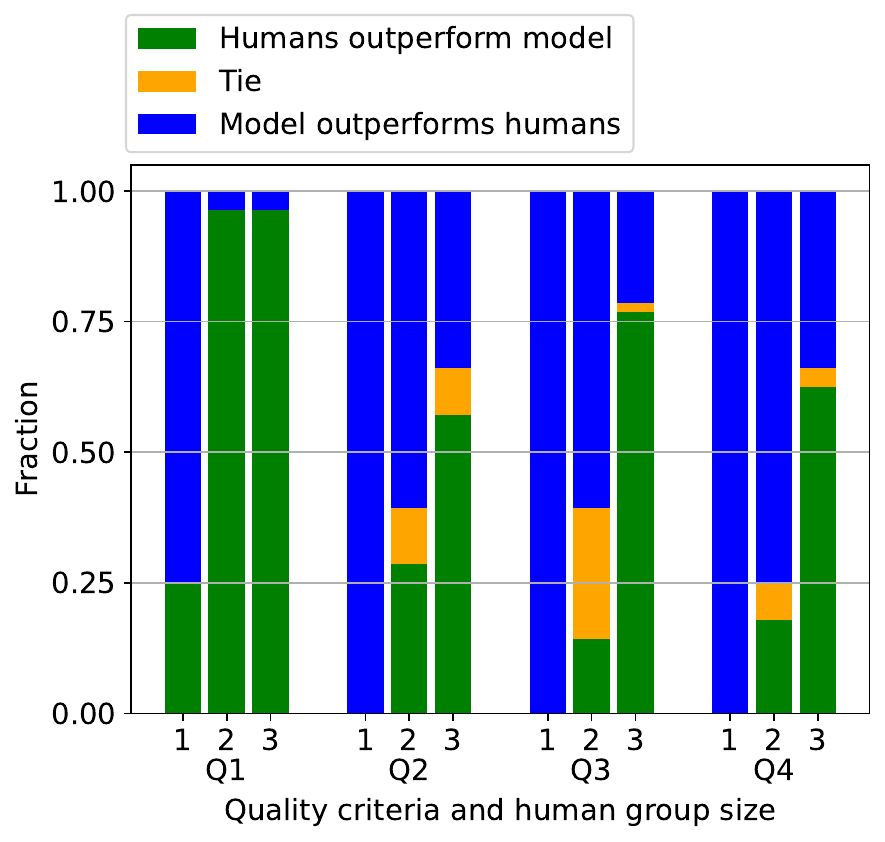}
\caption{Fraction of groups of humans outperformed by the model. The model `wins' if it rates more statements closer to the golden rating than a group of humans.}
\label{fig:humans-vs-model-8cg}
\end{figure}

We determine for each possible individual, pair and triplet whether the model or humans are closer to the golden score for each of the 30 contributions (there are 8-choose-g groups, where g is the group size). We report in Fig.~\ref{fig:humans-vs-model-8cg-times-30} the fraction of data points where the model outperforms the groups of human annotators. Groups of three humans always outperform the model for each question, while single humans never do. In groups of two, humans perform better on Q1, while the model outperforms humans on all other questions.
We also aggregate these scores over the 30 questions, and report in Fig.~\ref{fig:humans-vs-model-8cg} for how many groups the model rates more statements closer to the golden rating than the humans. The same observations apply as in Fig.~\ref{fig:humans-vs-model-8cg-times-30}, but the margins of victory are larger for both humans and the model.   
\ifthenelse{\boolean{includeappendix}}{
  An analogous Figure using the sum of distances from the golden rating on all 30 statements is provided in Appendix~\ref{app:sum-of-distance}.}{} 
 The takeaway from this experiment is that the model is a reasonable alternative to human annotators because three annotators would be too expensive for most deliberation events. 
\begin{figure}[h]
    \centering
   \includegraphics[trim=20pt 20pt 20pt 45pt, clip, scale = 0.55]{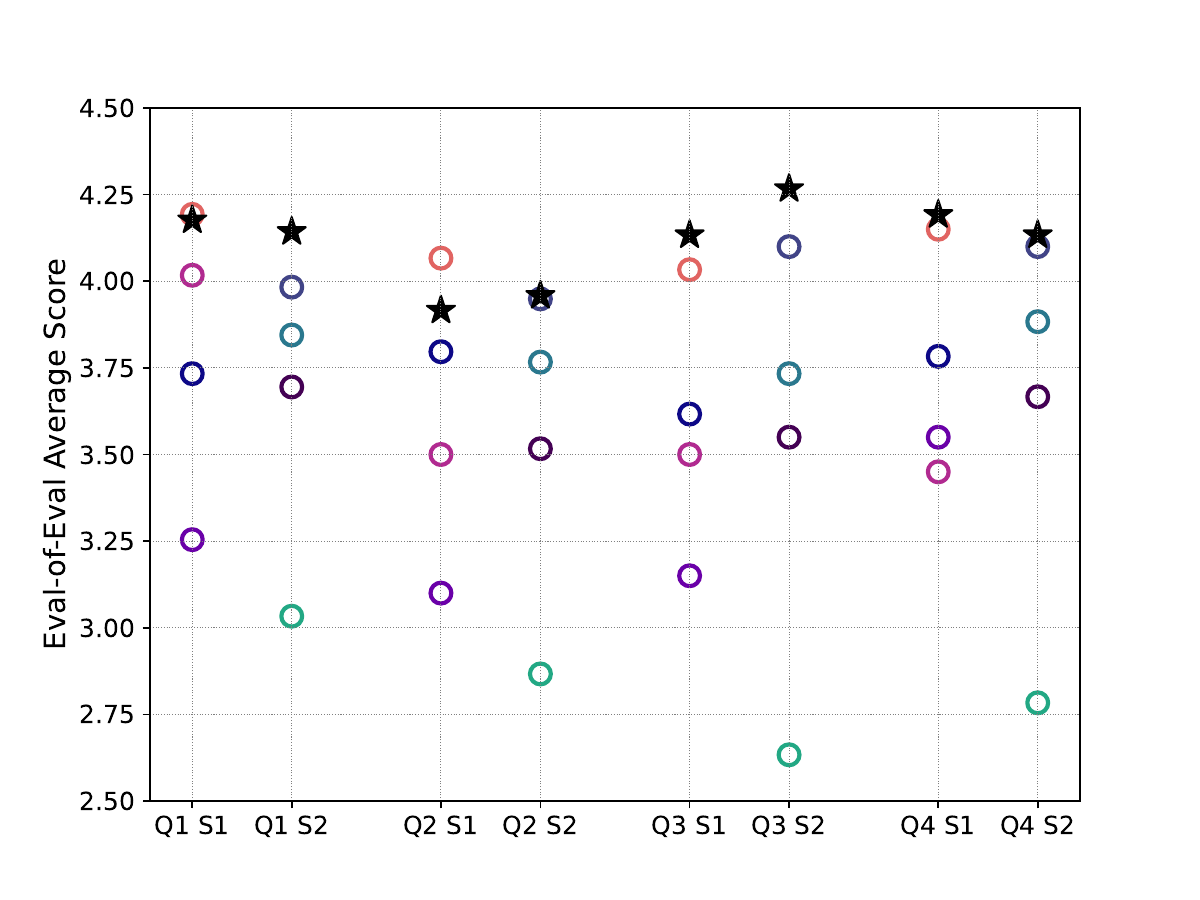}
\caption{ Average scores received in the evaluation of rating-justification pairs by individual annotators on the 1-5 Likert scale. S1 and S2 are the contribution sets. 
The model's results are marked by stars. Each annotator has a unique color used across all criteria. 
}
\label{fig:eval-of-rating}
\end{figure}
\begin{figure}[h]
    \centering
   \includegraphics[trim=12pt 25pt 15pt 50pt, clip, scale = 0.55]{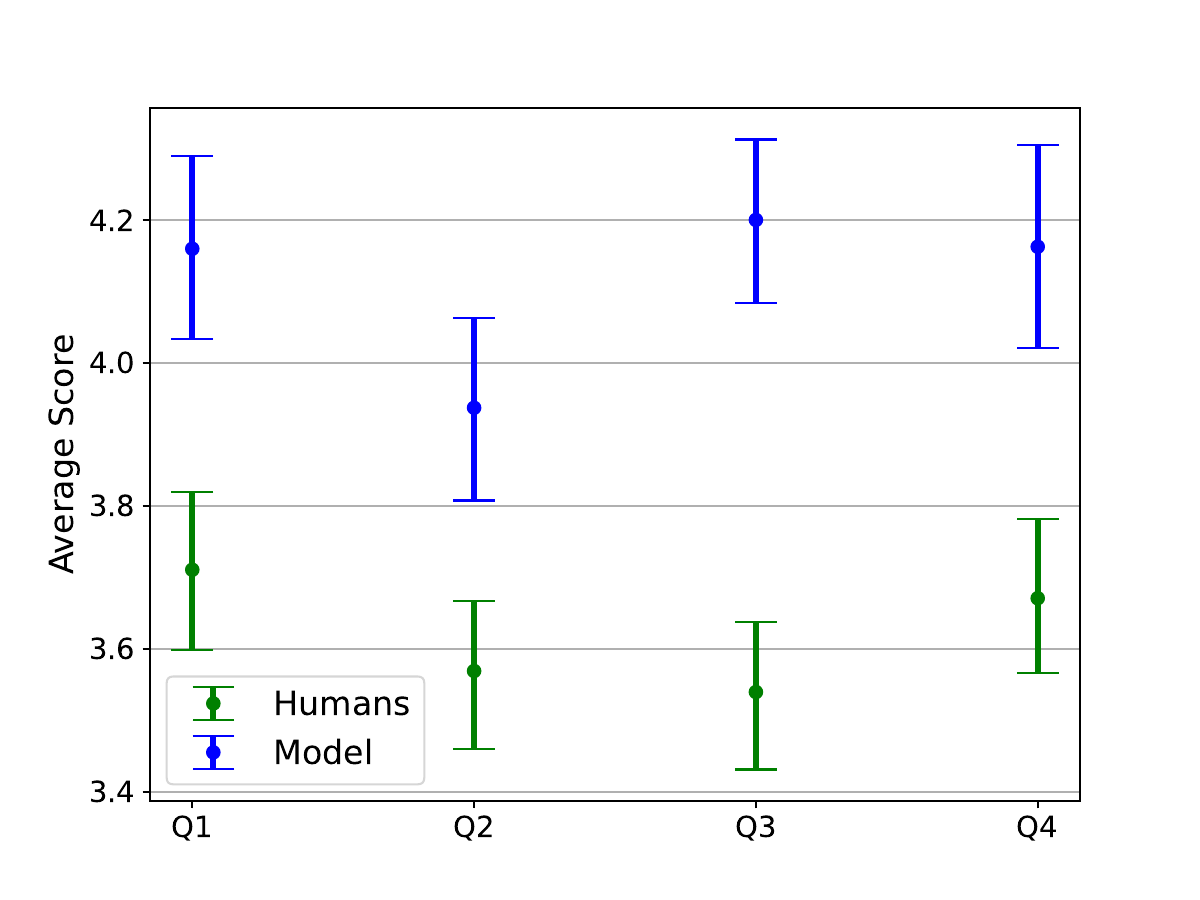}
\caption{Average scores received in evaluations of rating-justification pairs on the 1-5 Likert scale.}
\label{fig:avg-eval-of-ratings}
\end{figure}
\begin{figure}[h]
    \centering
    \includegraphics[scale = 0.55]{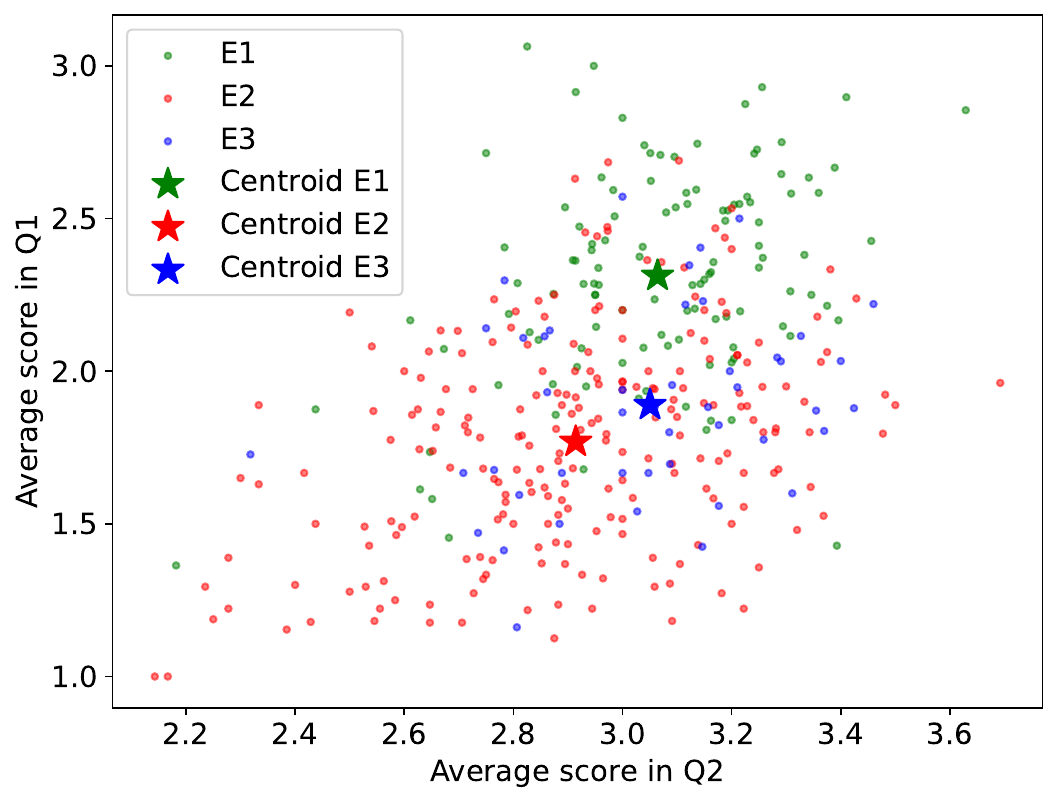}
    \caption{Average room quality on Q1 (justification) and Q2 (novelty) for all rooms in events E1, E2, and E3. The centroid corresponds to the average of the room quality for each room in the event. }
    \label{fig:avg-room}
\end{figure}
%
%
%
%
%
A second approach to evaluate the performance of the model against human annotators, is to ask other annotators to evaluate the scores and their justifications (See \S~\ref{sec:methodology evaluating}). We report the scores of both human and model ratings and justifications in Fig.~\ref{fig:eval-of-rating} and the averages in Fig.~\ref{fig:avg-eval-of-ratings}. The evaluators are not informed about the source of the ratings and justification. The model outperforms the humans by a significant margin, as is evident from the 95\% confidence intervals (obtained via bootstrapping by drawing 10,000 samples).
\ifthenelse{\boolean{includeappendix}}{
 The distribution of these scores is in Fig.~\ref{fig:dist-eval-of-ratings} in Appendix~\ref{app:dist_eval_of_eval}.}{}


\subsection{Efficiency}
The cost of annotating a 45-minute deliberation event with GPT-4 is under \$4. As of writing, OpenAI charges \$30 per million tokens input and \$60 per million tokens output. A typical 45-minute deliberation event contains 30-50 contributions, and we make four queries for each contribution, where a typical query has between 200-1000 tokens of input and 20 tokens of output. An additional advantage of automated quality estimation is that it can be implemented in real-time (under 2s delays for a query), allowing for platform interventions to be informed by it.

On the other hand, even if we assume a human annotator can consistently estimate quality in 20 seconds per criterion per contribution, at the California minimum wage of \$16 per hour, human annotations would be four times more expensive than this automated approach. Using three human annotators would triple the cost further. This doesn't even take into account the cost of training the human annotators.
%

\subsection{Analysis of Deliberation Events}
\label{sec:analysis of deliberations}
Since we have established that the model is a reliable annotator, we can use it at scale to inform platform design choices and estimate the average quality of all rooms in deliberation events. For example, in Fig.~\ref{fig:avg-room}, we plot the average quality of statements in rooms (average over all filtered contributions) for Q1 (justification) and Q2 (novelty). E1 has a higher overall quality on both questions than both E2 and E3 (both using the same agenda design). The figure also shows a large variation between rooms in the same event. This analysis can prove useful when evaluating the event to identify rooms that had a particularly meaningful conversation (for qualitative analysis) or when evaluating the event as a whole. 

\ifthenelse{\boolean{includeappendix}}{
 We report the average quality per agenda item for E1, E2, and E3 in Appendix~\ref{app:session_agenda_quality}. The variations in these results suggest potential guidance for designing future agenda items.}{} 

\section{Evaluating Nudges Via Quality Estimates}
\label{sec:nudges}
We can use the quality estimates to evaluate the effect of interventions on the quality of conversation. We will apply this to our speak nudges, which are emitted to a participant after three minutes of not requesting to speak (\S\ref{sec:methodology nudges}). We will first establish the effectiveness of the nudges towards increasing participation. We use dataset E4 here (recall \S\ref{sec:data}). 
We have data from 27,899 nudges \textit{sent} and 5,222 nudges \textit{skipped}. 

\subsection{Effectiveness at Prompting to Speak}

With the skipped nudges, we have a randomized controlled trial. We can calculate the probability that the participant requests to speak in the 30-second interval following the emission of a nudge (or a skipped nudge) and establish the effect of a nudge by taking the difference in means.
Out of the 27,899 nudges, 2,929 (10.4\%) resulted in a positive participation outcome, (a request to speak). Out of these, 2625 resulted in an actual contribution. 
The 5,222 instances where nudges were skipped were followed by 332 requests to speak (6.4\%) in 30 seconds, which then led to 301 statements.

\begin{figure}[h]
    \centering
   \includegraphics[trim=10pt 10pt 20pt 50pt, clip, scale = 0.55]{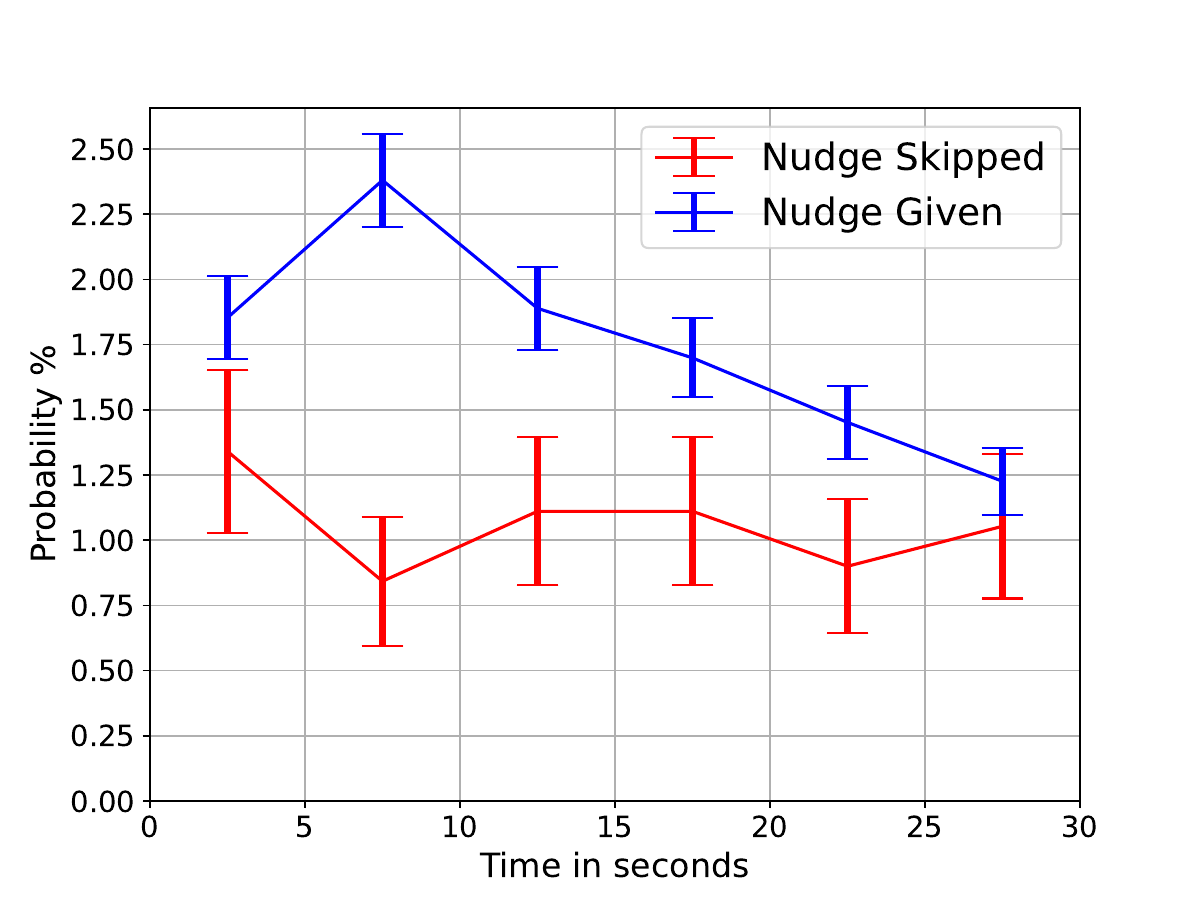}
\caption{Probability of a nudge leading to a speaking request within 30 seconds in dataset E4. Red bars show the control with skipped nudges, and blue bars show the case with nudges. Error bars represent 95\% confidence intervals for each 5-second interval.}
\label{fig:nudge_effect}
\end{figure}

This indicates that speak nudges result in a 65.1\% increase in the probability of a request to speak and a 63.2\% increase in the probability of a statement. In Fig.~\ref{fig:nudge_effect} we report these probabilities broken down in 5-second intervals, confirming the intuition that the effect of a nudge is most prominent a little after emission.

\ifthenelse{\boolean{includeappendix}}{
We report the effectiveness of repeated nudges in Fig.~\ref{fig:next_nudge} in Appendix~\ref{app:repeated_nudges}.}{We report the effectiveness of repeated nudges in the full version of the paper.} 
 The first nudge is the most effective (around 15\%), with effectiveness dropping quickly. By the fourth nudge, the effectiveness is comparable to a skipped nudge. This shows that sending many nudges is unproductive and that fewer, strategically timed nudges is better. 
\subsection{Effect of Nudges on Contribution Quality}
Now that we have established that nudges increase participation, we are interested in whether those statements are higher or lower quality.
We study the effect of nudging on contribution quality in three ways:
\begin{description}[noitemsep]
    \item [Analysis 1.] Compare the average quality of contributions following emitted and skipped nudges in dataset E4.
    \item [Analysis 2.] Compare the average quality of contributions after nudges to all other contributions in E1, E2, and E3.
    \item [Analysis 3.] Compare the average quality of contributions after nudges to all other contributions for \textit{individual} participants in datasets E1, E2, and E3.
\end{description}
We say a contribution follows a nudge (or a skipped nudge) if the speaking request comes in the following 30 seconds.

\begin{figure}[h]
    \centering
    \includegraphics[scale = 0.55]{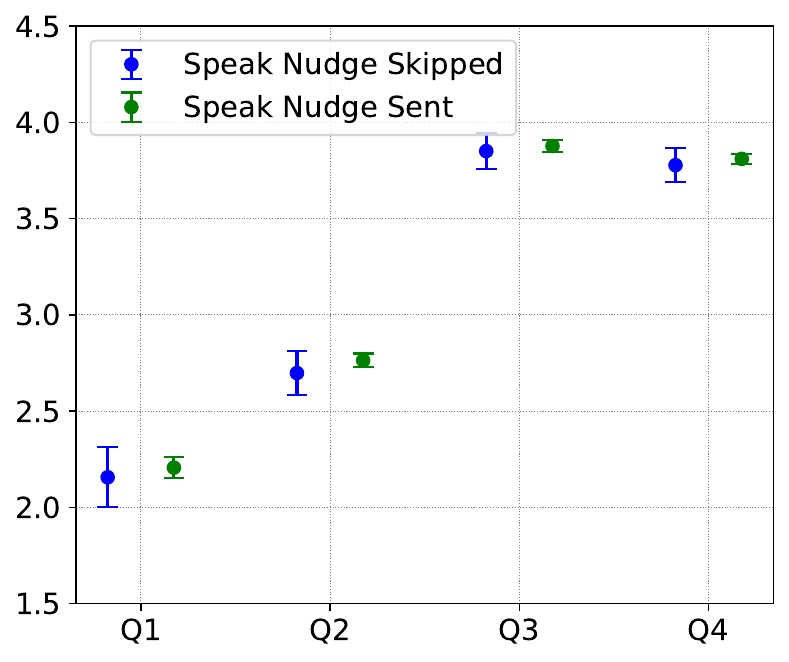}
    \caption{Average quality of statements for which the  associated speaking request came after a skipped nudge or a sent nudge with 95\% confidence intervals in dataset E4.}
    \label{fig:skip_nudge_vs_nudge}
\end{figure}

\paragraph{Analysis 1.}
The average quality score for contributions following a nudge is not less than that for contributions following skipped nudges. These are higher by $0.05, 0.07, 0.03,$ and $0.03$ on Q1, Q2, Q3, and Q4, respectively. See  Fig.~\ref{fig:skip_nudge_vs_nudge} for 95\% confidence intervals of the scores. Recall that nudges are sent (or randomly skipped) to participants with lower activity. This result is an important point in favour of targeted nudges. While nudges increase engagement, they do so without decreasing the quality of contributions, thereby generating more ideas in the conversation.  



\begin{figure}[h]
    \centering
    \includegraphics[scale = 0.55]{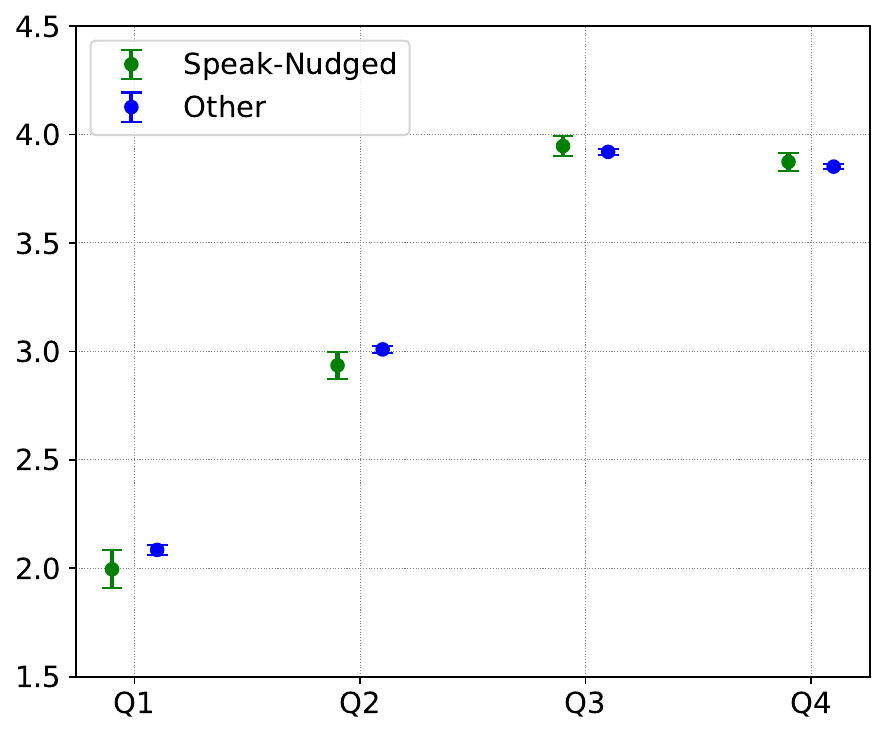}
    \caption{Average quality of nudged and other statements in events E1, E2, and E3, with 95\% confidence intervals.}
    \label{fig:all_datasets_quality_vs_nudge_or_no_nudge}
\end{figure}

\paragraph{Analysis 2.}
When compared with all other contributions, we observe that the average quality of nudged contributions is lower on justification (Q1) and novelty (Q2) and higher on constructiveness (Q3), and allowing expansion (Q4) (Fig.~\ref{fig:all_datasets_quality_vs_nudge_or_no_nudge}). 
This difference in average quality is $-0.09, -0.07, 0.03,$ and $0.02$ for Q1, Q2, Q3, and Q4, respectively. This is small compared to the standard deviation of the scores, which are $1.36, 0.96, 0.79,$ and $0.70$ in the same order. 
%
 This result does not imply causation since nudges are more likely to be sent to less active participants.
\ifthenelse{\boolean{includeappendix}}{We give further breakdowns of these results across events and genders in Appendix~\ref{app:nudged_quality}.
}{We give further breakdowns of these results across events and genders in the full version.}

\paragraph{Analysis 3.}
For each participant, we calculated the difference in the average quality of their nudged and other contributions. We then averaged this difference across all participants with at least one nudged and one non-nudged contribution. This is 607 participants over E1, E2, and E3. The results are shown in
(Table~\ref{tab:individual_nudge_effect}). The small negative effect seen in Analysis 2 is now reduced to near zero!

\begin{table}[h]
    \centering
    \begin{tabular}{p{0.5cm} >{\centering\arraybackslash}m{3cm} >{\centering\arraybackslash}m{3.5cm}}
        \toprule
        Q & Nudge Effect & Nudge Effect 95\% CI  \\
        \midrule
        Q1 & -0.012 & [-0.112, 0.089] \\
        Q2 & 0.022 & [-0.046, 0.090] \\
        Q3 & -0.008  & [-0.062, 0.043]  \\
        Q4 & 0.021  & [-0.024, 0.065]  \\
        \bottomrule
    \end{tabular}
    \caption{ Nudge effect on individual-level quality, averaged over all participants in events E1, E2, and E3. Negative numbers indicate that the quality is lower for nudged statements. }
    \label{tab:individual_nudge_effect}
\end{table}

The intuition that the rating difference between nudged and other contributions is driven by the fact that nudges are sent to less-active individuals is further corroborated by the following observation. The Pearson Correlation Coefficient between the number of contributions made by a participant and their average quality scores are:
Q1: 0.21, Q2: 0.24, Q3: 0.13, and Q4: 0.17, all nontrivial positive numbers, indicating a weak correlation between the average activity level and the quality of contributions. 
Analysis 3 further reinforces the takeaway of Analysis 1 that nudged statements have the same average quality as non-nudged statements.

\section{Conclusions and Future Work}
\label{sec:discussion}

In this work, we demonstrate that LLMs are valuable tools for rating deliberative contributions, providing consistent and scalable quality assessments that significantly reduce the resource demands associated with human annotations. This approach aids in evaluating platform interventions, such as targeted nudges, and overall deliberation quality.  

One significant limitation is the unpredictability of LLMs \cite{perkovic2024hallucinations}. In our tests on 30 randomly selected statements across 5 trials, we observed low average variances of 0.13, 0.08, 0.02, and 0.01 in the model's ratings on Q1-Q4, respectively, indicating good consistency in the model’s ratings. Ensuring this consistency across different contexts remains crucial. Additionally, using AI in this context raises issues of transparency and accountability. LLM outputs are highly sensitive to prompt design \cite{sclar2023quantifying, errica2024did}, and the rapid development of LLMs presents the risk that current queries may not work as expected with future models. Analysts must remain vigilant about potential biases and may need to adjust queries over time. Evaluating AI-generated justifications can aid explainability, but ongoing human quality control may still be necessary. While developing a self-hosted LLM could mitigate some risks, it would also introduce additional costs and complexity. 

Further research includes improving the model through fine-tuning and mean correction to enhance the accuracy of quality assessments. Another key direction is studying the relationship between discussion quality and opinion change, measured through pre- and post-event surveys, to understand how discourse quality influences participant perspectives. Designing platform interventions based on real-time quality assessments is an exciting avenue for future work, potentially allowing dynamic adjustments to enhance the effectiveness of deliberative processes. Integrating automated assessments with other evaluation metrics could provide a more comprehensive understanding of deliberative quality.

%% file: appendix.tex

\appendix
\section{Appendix}
\label{sec:reference_examples}

\subsection{Distribution of Ratings Given by the Model and Humans}
\label{app:scores}
\begin{figure}[h]
    \centering
    \includegraphics[scale = 0.5]{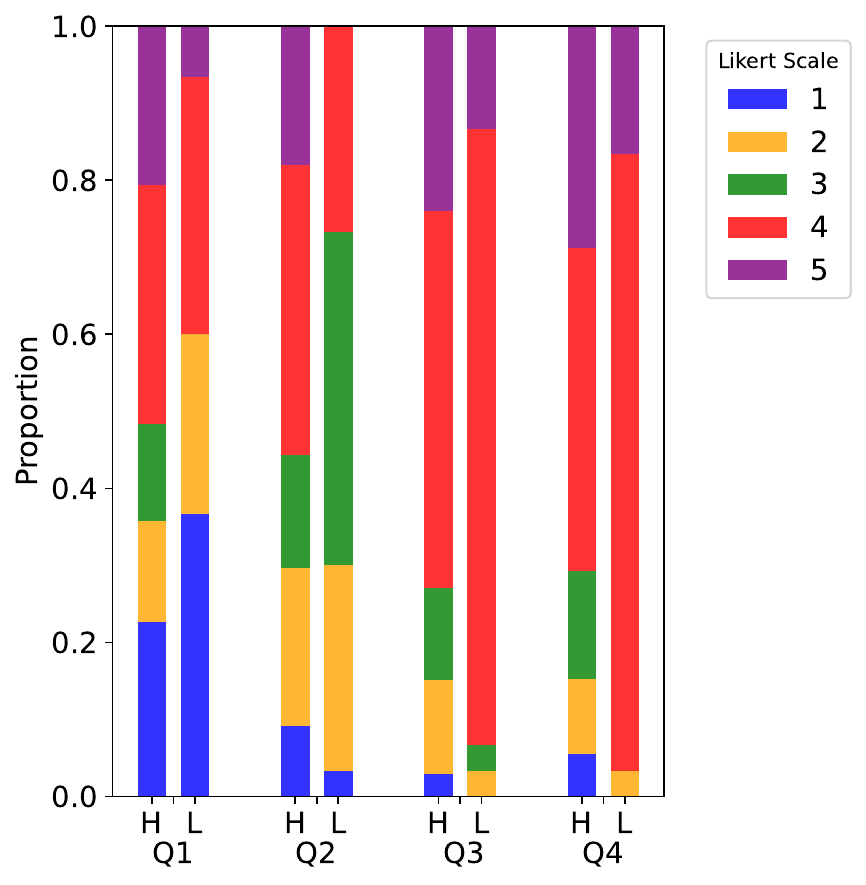}
    \caption{The ratings given by the model and humans across 30 statements. H: Human; L: Language model.}
    \label{fig:scores-bar}
\end{figure}

\subsection{Effectiveness of Subsequent Nudges}
\label{app:repeated_nudges}

\begin{figure}[h]
    \centering
   \includegraphics[scale = 0.5]{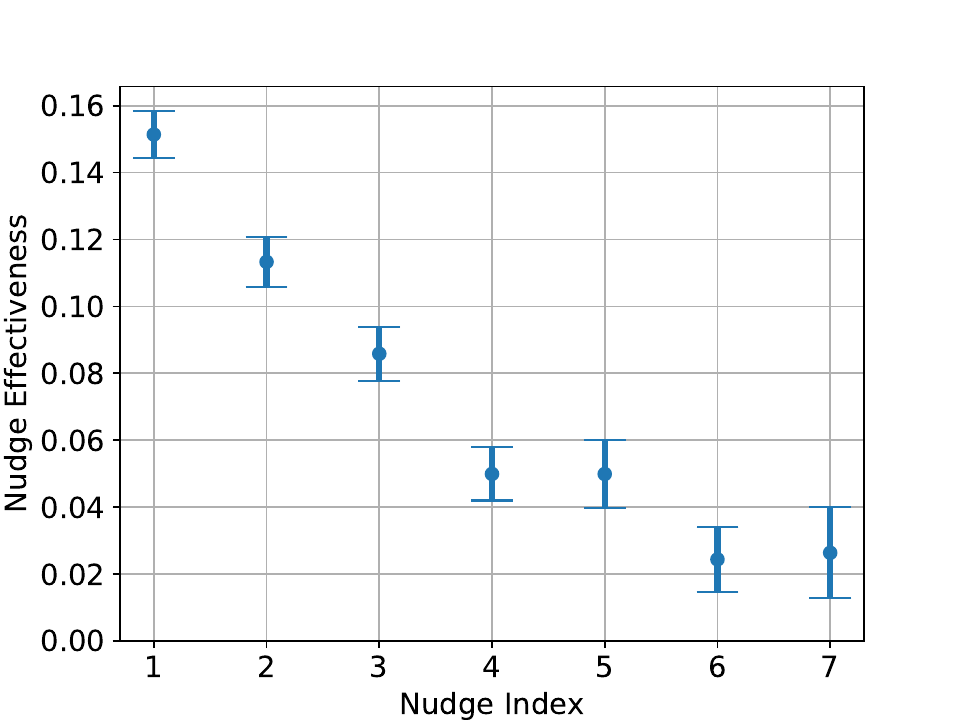}
\caption{Effectiveness of subsequent nudges measured as probability of the participant making a speaking request in the next 30 seconds. The confidence intervals are for 95\%.}
\label{fig:next_nudge}
\end{figure}

\newpage
\subsection{Quality of Nudged and Other Contributions For Different Genders}
\label{app:nudged_quality}

\begin{figure}[h]
    \centering
    \includegraphics[scale = 0.44]{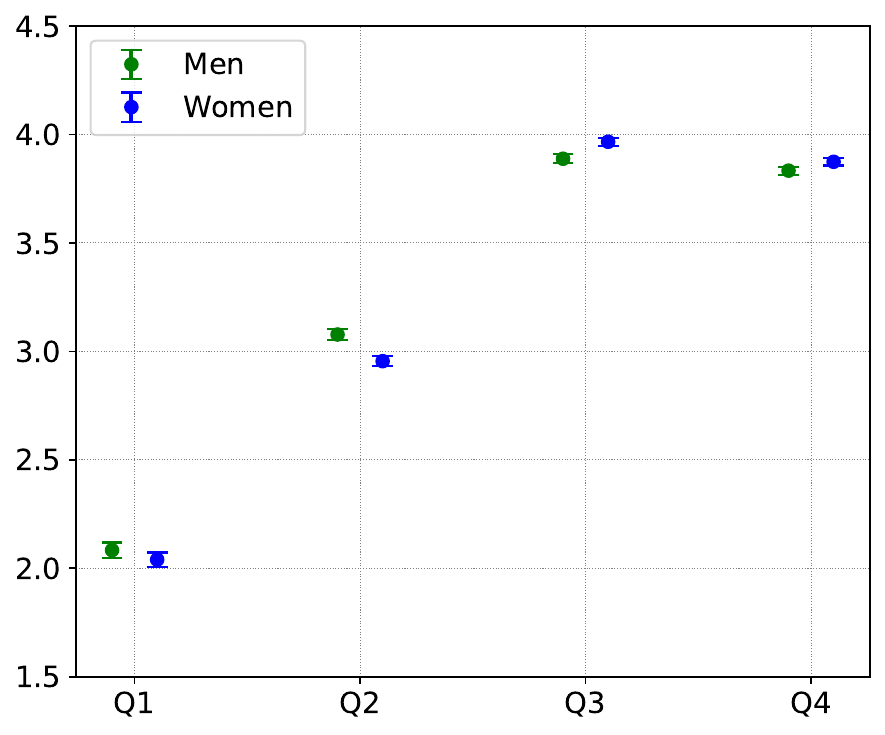}
    \caption{Average quality for men and women across Events E1, E2, and E3 with 95\% CI. }
    \label{fig:enter-label}
\end{figure}

\begin{figure}[h]
    \centering
    \includegraphics[scale = 0.44]{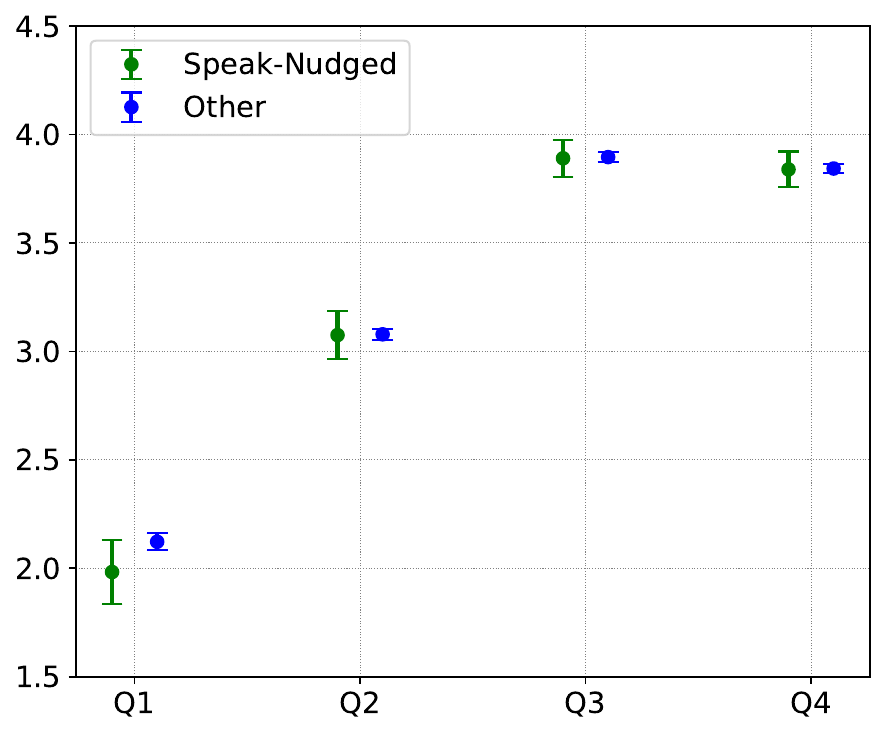}
    \caption{Average quality of nudged and other statements across Events E1, E2, and E3 for men, with 95\% CI.}
    \label{fig:enter-label}
\end{figure}

\begin{figure}[h]
    \centering
\includegraphics[scale = 0.44]{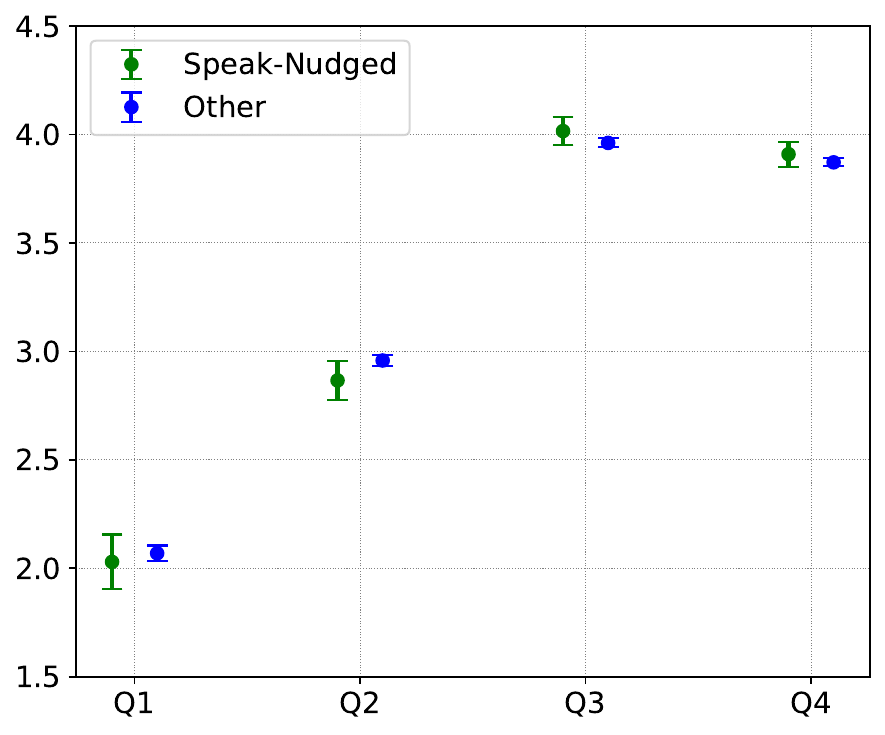}
    \caption{Average quality of nudged and other statements across Events E1, E2, and E3 for women, with 95\% CI.}
    \label{fig:enter-label}
\end{figure}

\newpage
\subsection{Quality of Nudged and Other Contributions For Different Events}

\begin{figure}[h]
    \centering
    \includegraphics[scale = 0.44]{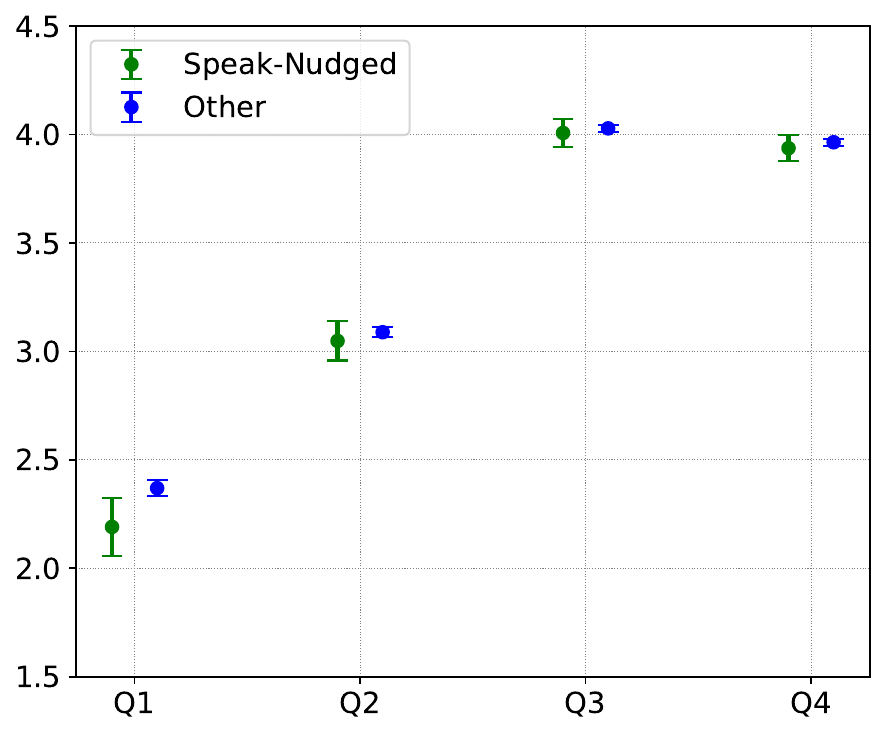}
    \caption{Average quality of nudged and other statements in Event E1, with 95\% CI.}
    \label{fig:enter-label}
\end{figure}

\begin{figure}[h]
    \centering
    \includegraphics[scale = 0.44]{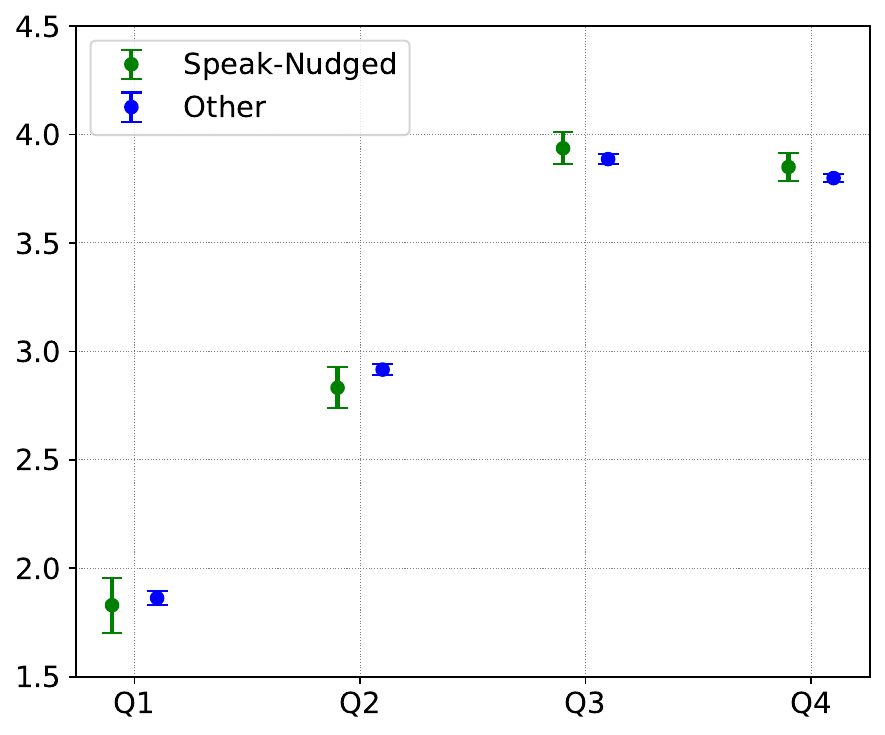}
    \caption{Average quality of nudged and other statements in Event E2, with 95\% CI.}
    \label{fig:enter-label}
\end{figure}

\begin{figure}[h]
    \centering
    \includegraphics[scale = 0.44]{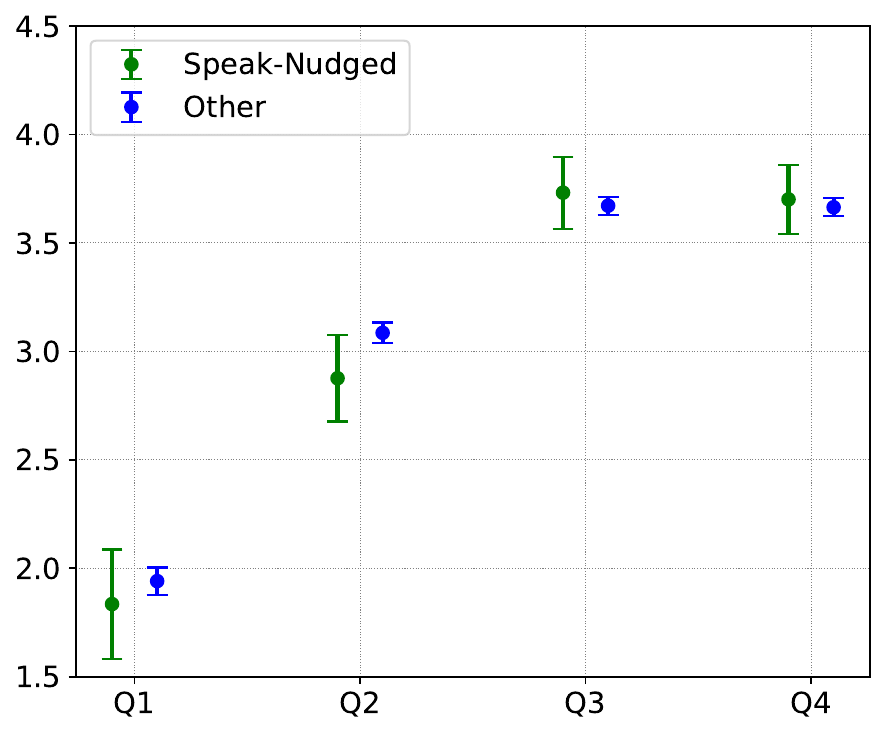}
    \caption{Average quality of nudged and other contributions in Event E3, with 95\% CI.}
    \label{fig:enter-label}
\end{figure}

\newpage
\subsection{Quality of Contributions of Men and Women For Different Events}
\begin{figure}[h]
    \centering
    \includegraphics[scale = 0.44]{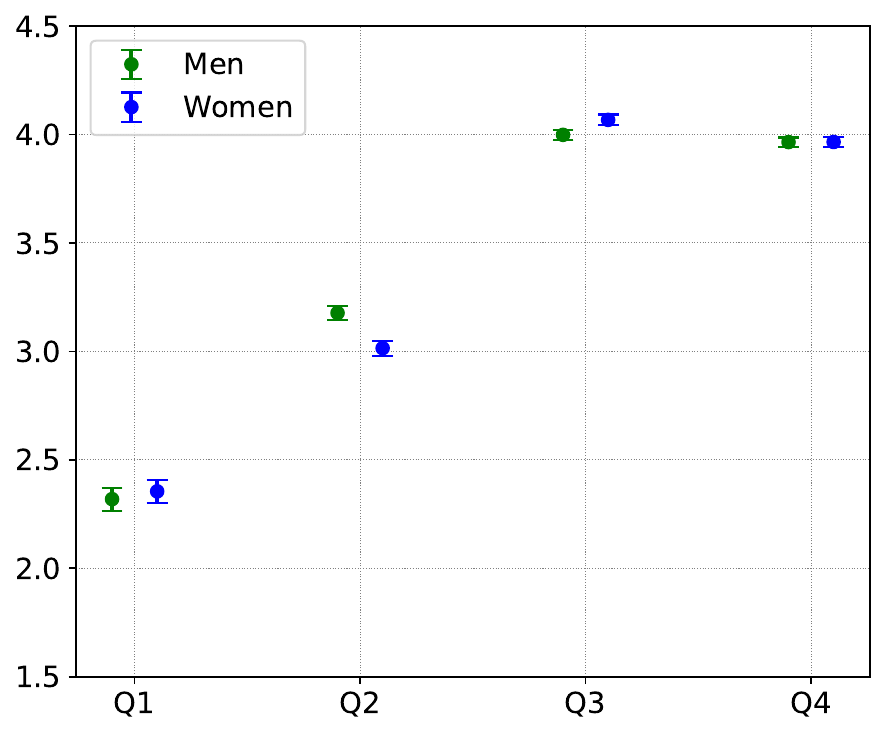}
    \caption{Average quality of contributions by men and women in Event E1 with 95\% CI.}
    \label{fig:enter-label}
\end{figure}
\begin{figure}[h]
    \centering
    \includegraphics[scale = 0.44]{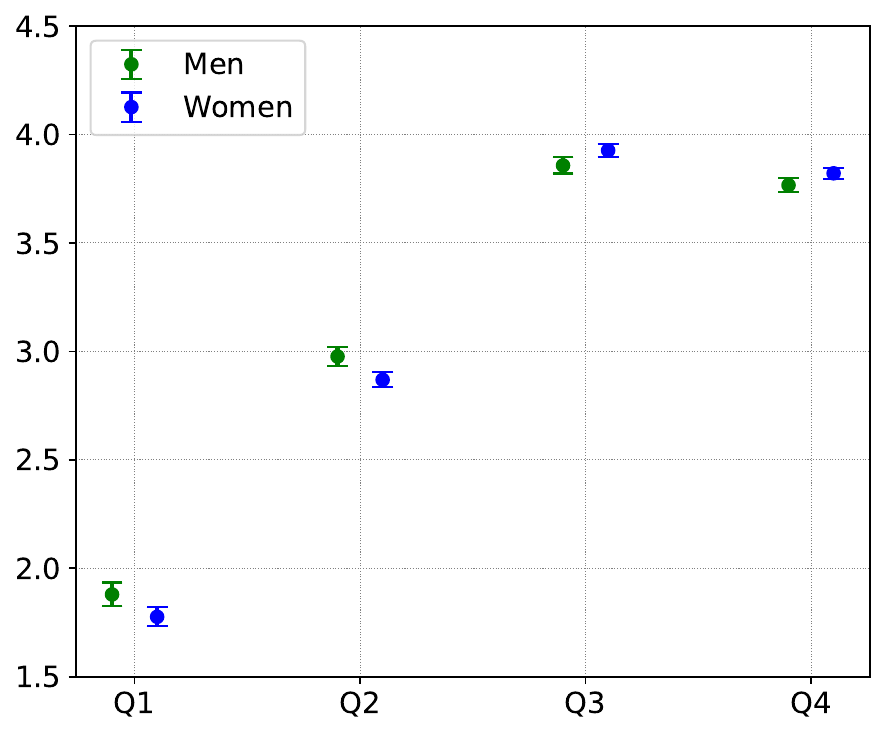}
    \caption{Average quality of contributions by men and women in Event E2 with 95\% CI.}
    \label{fig:enter-label}
\end{figure}
\begin{figure}[h]
    \centering
    \includegraphics[scale = 0.44]{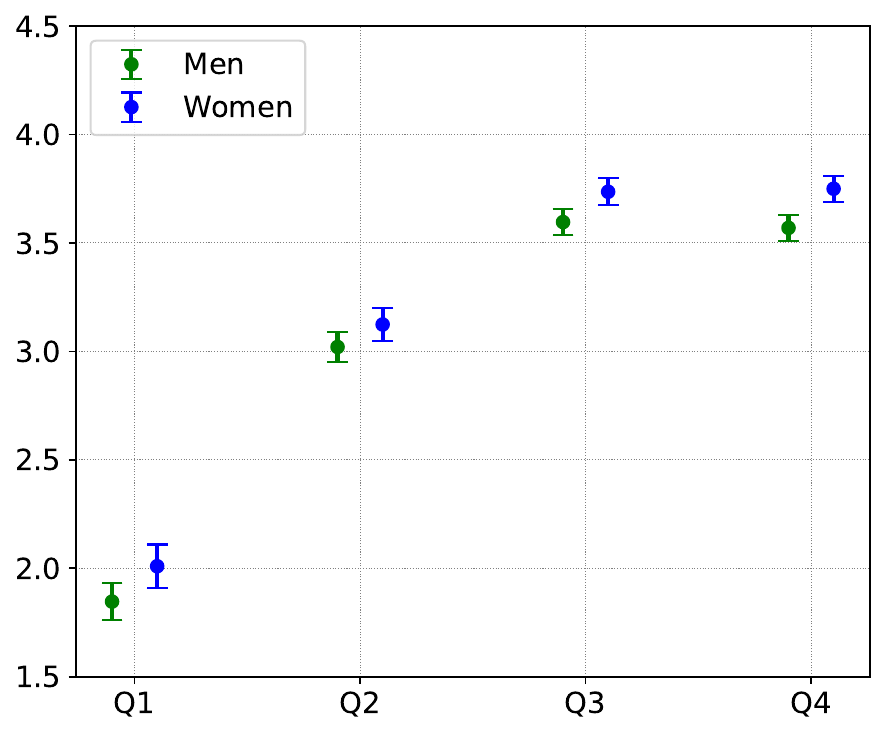}
    \caption{Average quality of contributions by men and women in Event E3 with 95\% CI.}
    \label{fig:enter-label}
\end{figure}
\newpage
\subsection{Distribution of Scores in Evaluation of Rating-Justification Pairs}
\label{app:dist_eval_of_eval}
\begin{figure}[h]
    \centering
    \includegraphics[trim=2pt 2pt 5pt 5pt, clip, scale = 0.45]{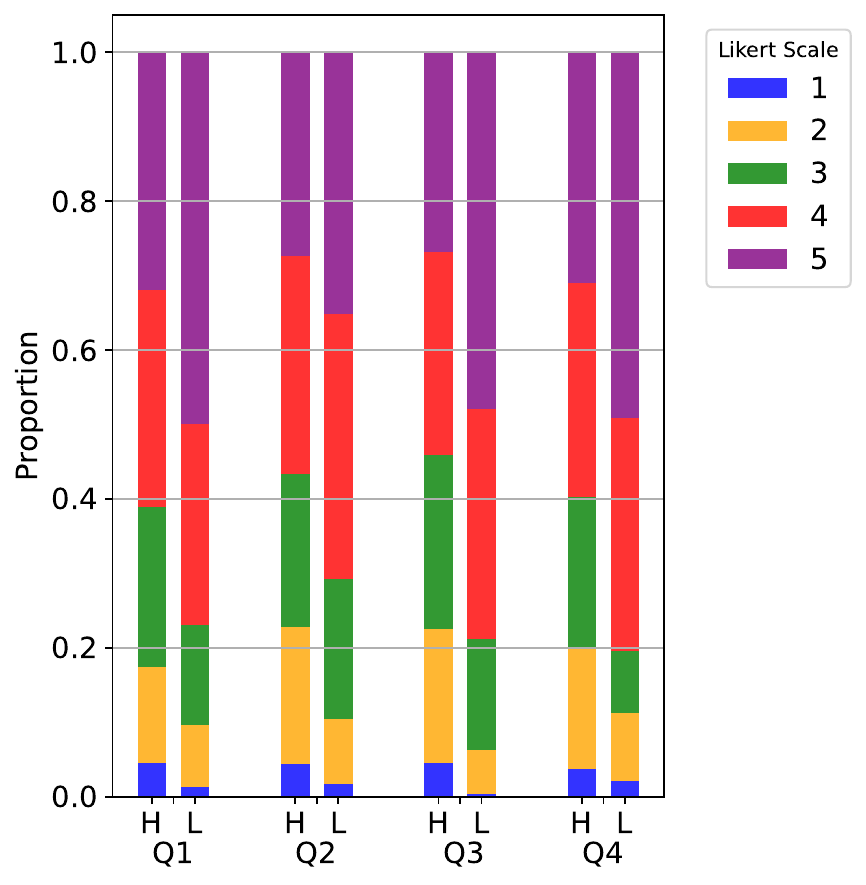}
    \caption{Scores received by the model and humans in the evaluation of rating-justification pairs. H refers to humans, and L refers to the language model. The model consistently receives better scores than humans, with a score of at least 4 at least 70\% of the time.}
    \label{fig:dist-eval-of-ratings}
\end{figure}

\subsection{Groups of Humans Versus the Model}
\label{app:sum-of-distance}
\begin{figure}[h]
    \centering
   \includegraphics[scale = 0.5]{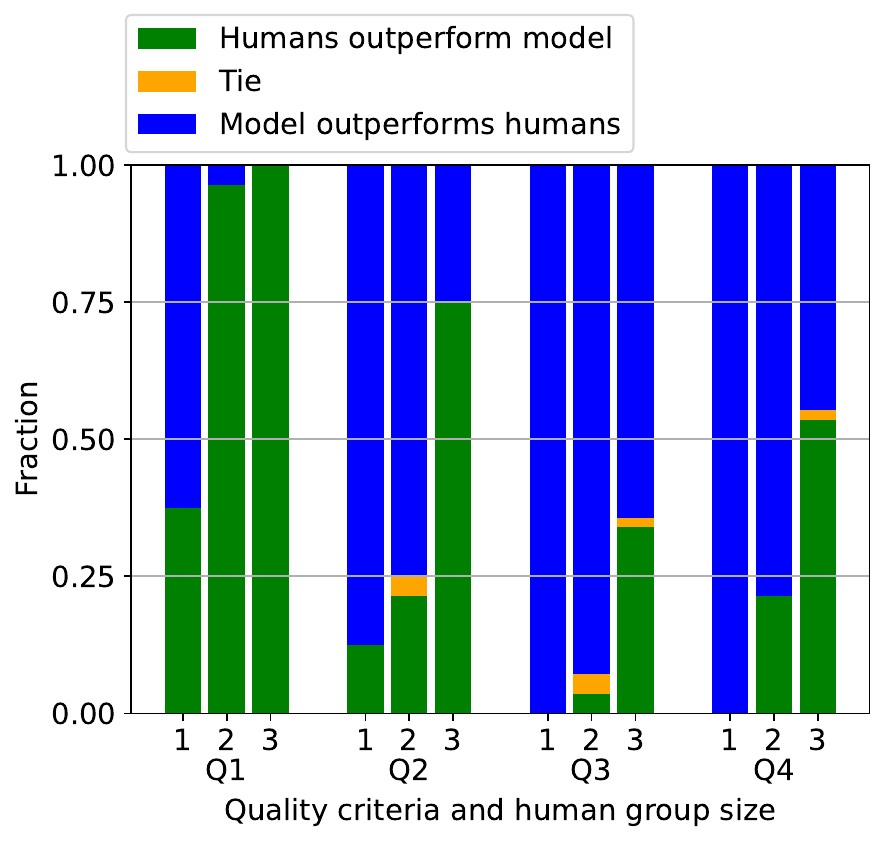}
   \caption{Fraction of groups of humans that are outperformed by the model. The model `wins' if its 1-norm across statements is smaller than a group of humans.}
\label{fig:humans_vs_model_distance}
\end{figure}

\newpage
\subsection{Average Quality of Statements in Different Sessions and Agenda Items}
\label{app:session_agenda_quality}

These results indicate that the quality does not vary significantly over time. It, however, depends on the specific details of the agenda items, giving us guidelines for the future design of agenda items.

\begin{figure}[h]
    \centering
    \includegraphics[scale = 0.4]{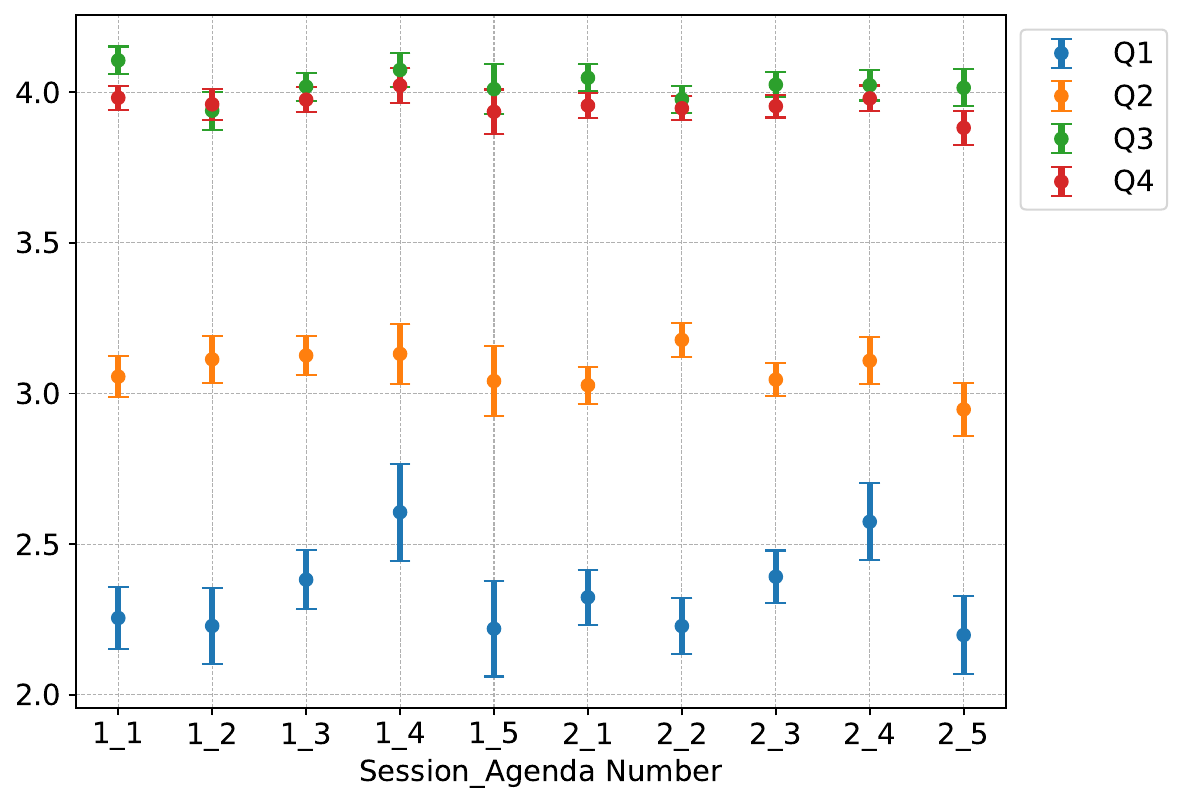}
    \caption{Average quality of contributions across the session and agenda numbers in Event E1.}
    \label{fig:enter-label}
\end{figure}

\begin{figure}[h]
    \centering
    \includegraphics[scale = 0.4]{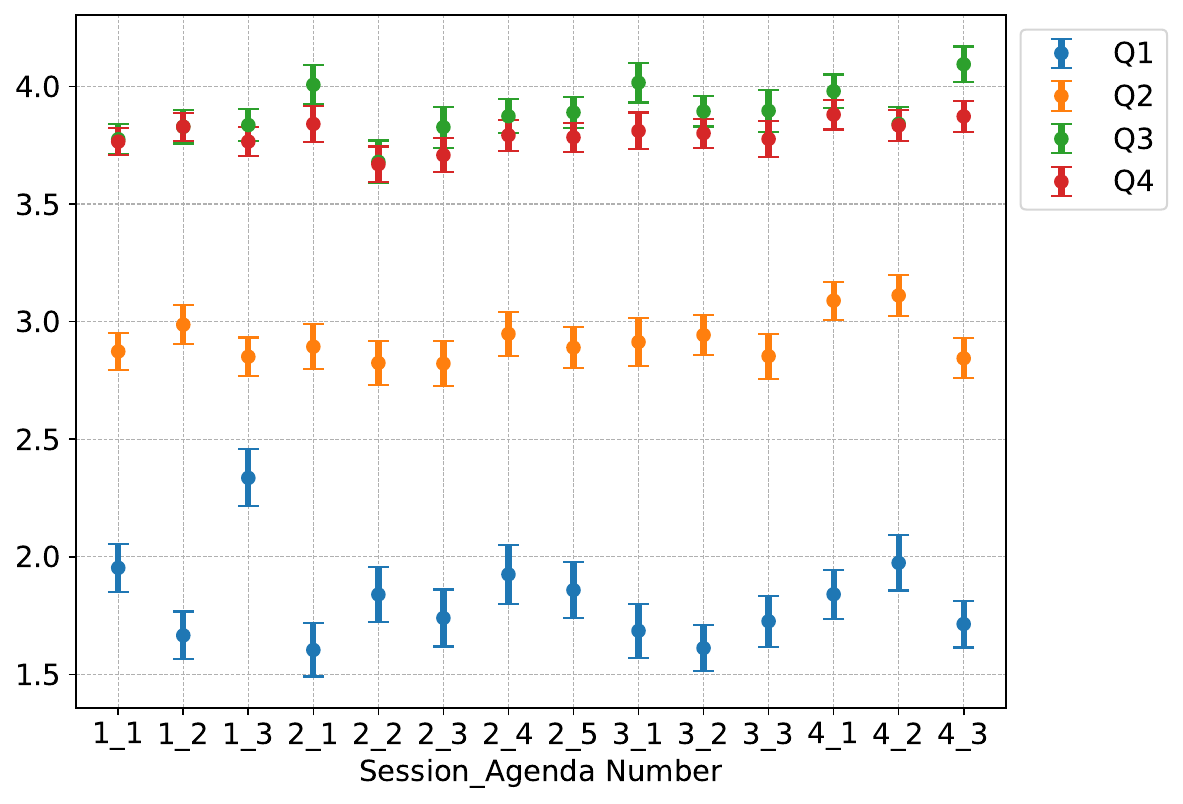}
    \caption{Average quality of contributions across the session and agenda numbers in Event E2.}
    \label{fig:enter-label}
\end{figure}

\begin{figure}[h]
    \centering
    \includegraphics[scale = 0.4]{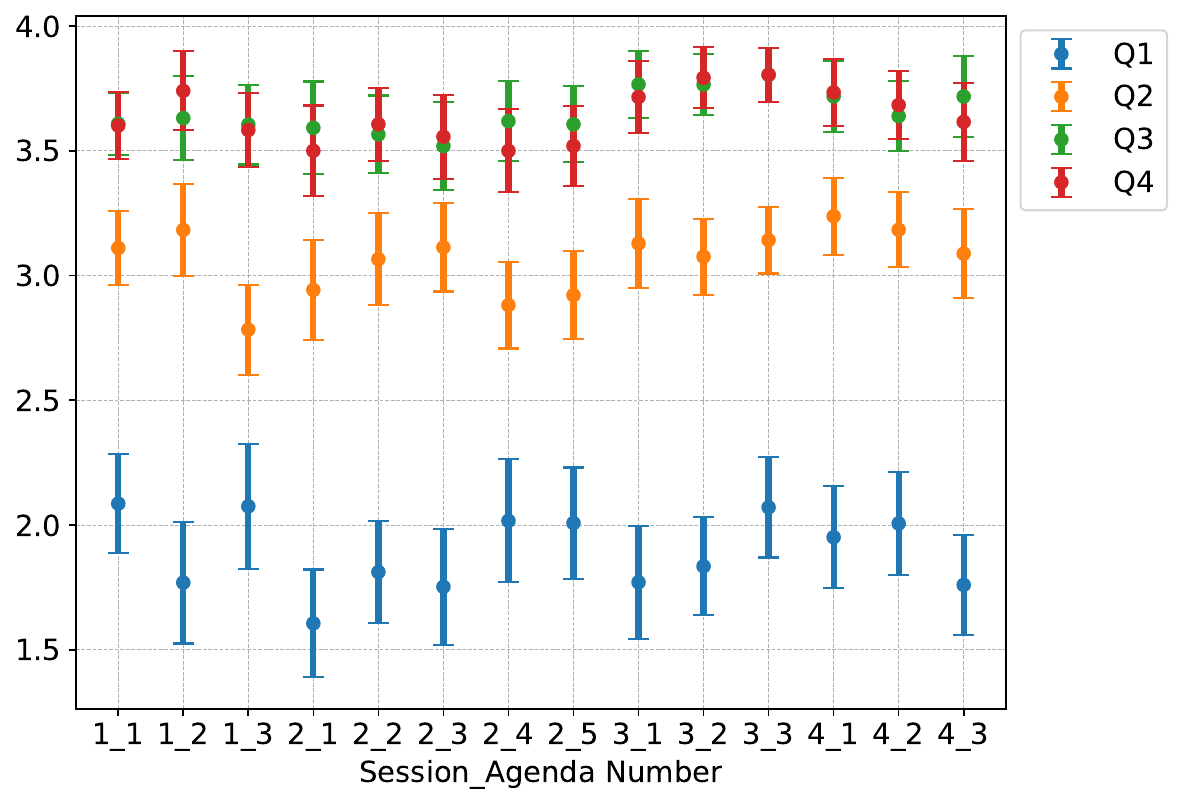}
    \caption{Average quality of contributions across the session and agenda numbers in Event E3.}
    \label{fig:enter-label}
\end{figure}

\newpage
\subsection{Improvement in the Model's Performance with Mean Correction}
\label{appendix-mean correction}
\begin{figure}[h]
    \centering
    \includegraphics[scale = 0.5]{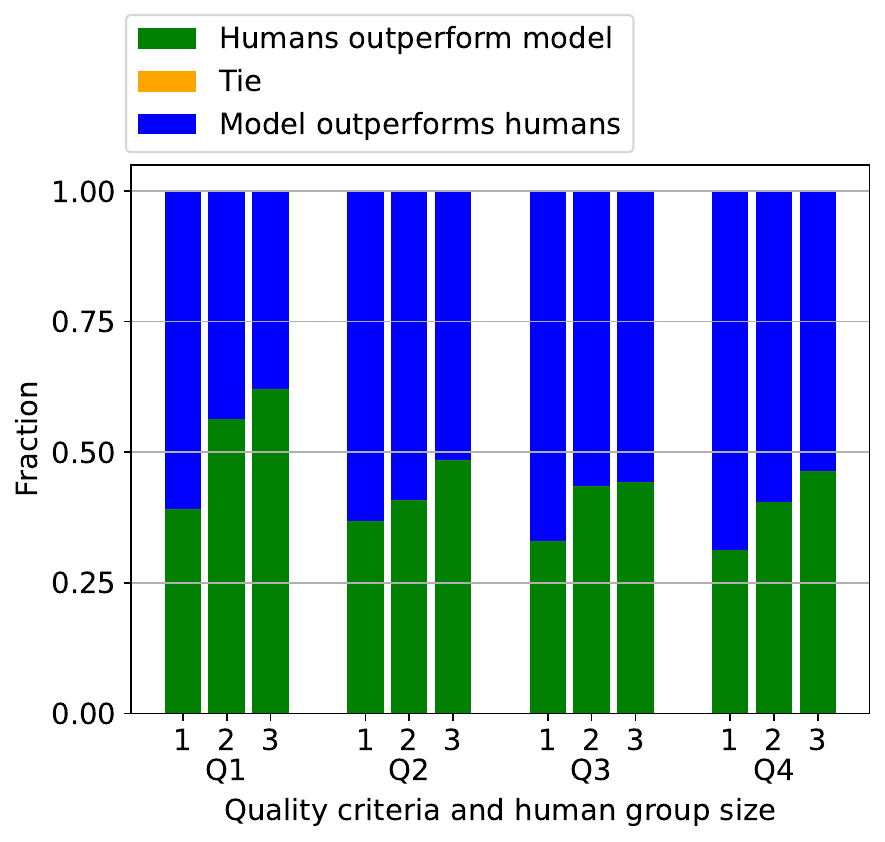}
    \caption{Fraction of data points where groups of humans are outperformed by the debiased model. The model `wins' if it rates a statement closer to the golden rating than a group of humans. Observe that the debiased model's performance is better than the original model in Fig~\ref{fig:humans-vs-model-8cg-times-30} on all four criteria.}
    \label{fig:debiased1}
\end{figure}

\begin{figure}[h]
    \centering
    \includegraphics[scale = 0.5]{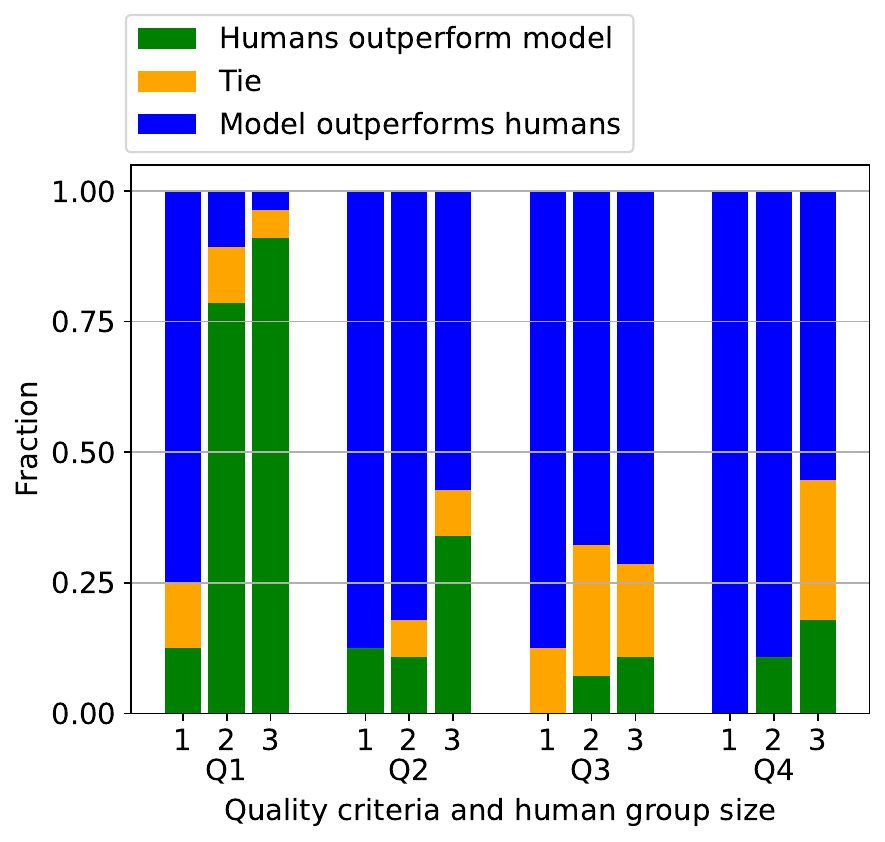}
    \caption{Fraction of data points where groups of humans are outperformed by the debiased model. The debiased model `wins' if it rates a statement closer to the golden rating than a group of humans. Observe that the debiased model's performance is better than the original model in Fig~\ref{fig:humans-vs-model-8cg} on all four criteria.}
    \label{fig:debiased2}
\end{figure}

\begin{figure}[h]
    \centering
    \includegraphics[scale = 0.5]{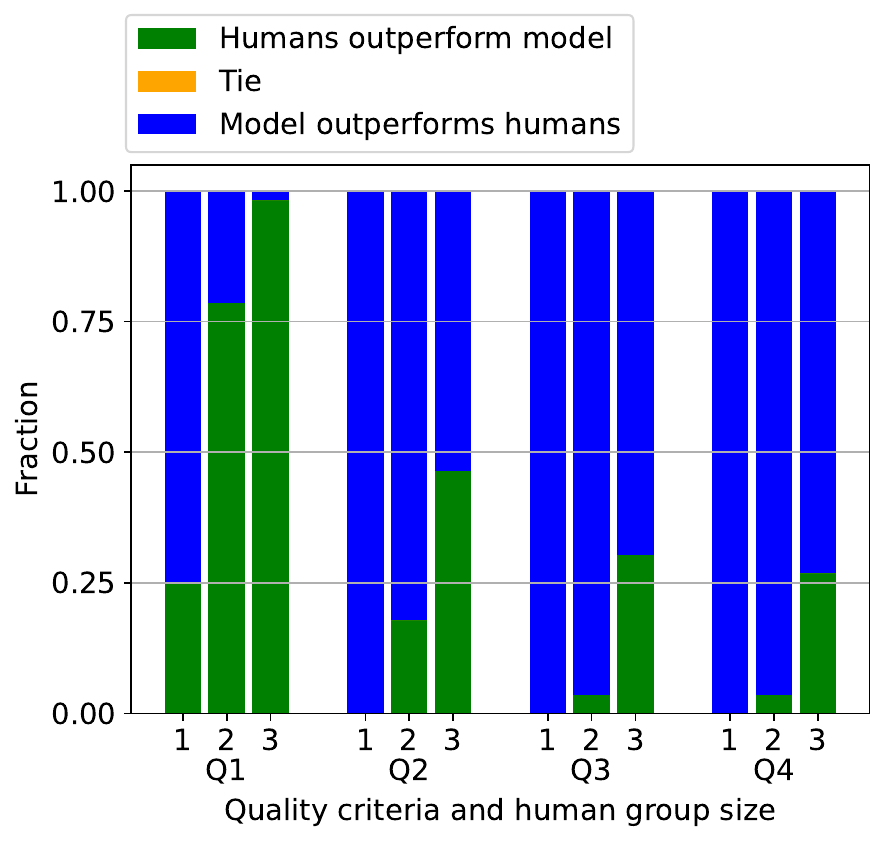}
    \caption{Fraction of groups of humans that are outperformed by the debiased model. The debiased model `wins' if its 1-norm across statements is smaller than a group of humans. Observe that the debiased model's performance is better than the original model in Fig~\ref{fig:humans_vs_model_distance} on all four criteria.}
    \label{fig:debiased3}
\end{figure}

In this analysis, we created a debiased rating for the model for each statement by subtracting the bias observed on the other 29 statements. We then used this debiased rating to compete with the human annotators. The results are in Figs.~\ref{fig:debiased1},~\ref{fig:debiased2},~\ref{fig:debiased3}. The model's performance improves on all quality criteria and all group sizes. Notably, it is now able to beat groups of three humans on all but one quality criterion.

